\documentclass{article} % For LaTeX2e
\usepackage{iclr2022_conference,times}

\usepackage{amsmath,amsfonts,bm}

\def\eqref#1{equation~\ref{#1}}
\def\1{\bm{1}}

\DeclareMathAlphabet{\mathsfit}{\encodingdefault}{\sfdefault}{m}{sl}
\SetMathAlphabet{\mathsfit}{bold}{\encodingdefault}{\sfdefault}{bx}{n}

\usepackage{hyperref}
\usepackage{url}

\hypersetup{
    colorlinks=true,
    linkcolor=red,
    urlcolor=blue,
    citecolor=darkgray
}
\usepackage{graphicx}
\usepackage{amsmath}
\usepackage{amssymb}
\usepackage{enumerate}
\usepackage{booktabs}       % professional-quality tables
\usepackage{multirow}
\usepackage{pifont}% http://ctan.org/pkg/pifont
\usepackage{xcolor}
\usepackage{xspace}
\usepackage{wrapfig}
\definecolor{TAR}{HTML}{bf9000}
\definecolor{REF}{HTML}{1155cc}
\definecolor{LIDAR}{HTML}{e04491}
\definecolor{Highlight}{HTML}{39b54a}
\usepackage{array}
\newcolumntype{H}{>{\setbox0=\hbox\bgroup}c<{\egroup}@{}}
\newcommand{\cmark}{\ding{51}}%
\newcommand{\xmark}{\ding{55}}%

\renewcommand{\paragraph}[1]{{\textbf{#1}}}
\makeatletter
\DeclareRobustCommand\onedot{\futurelet\@let@token\@onedot}
\def\@onedot{\ifx\@let@token.\else.\null\fi\xspace}

\def\eg{\emph{e.g}\onedot} 
\def\ie{\emph{i.e}\onedot} 
 
\def\etc{\emph{etc}\onedot}

\makeatother

\newcommand{\ourproposedmethod}{R4D}
\newcommand{\Ourproposedmethod}{R4D}
\newcommand{\longrangdataset}{anonymous internal long-range}

\newcommand{\LongRangDataset}{Anonymous Internal Long-range}
\newcommand{\lidar}{LiDAR}

\newcommand{\lowerbetter}{\color{red} \mathbf{\downarrow}}
\newcommand{\higherbetter}{\color{red} \mathbf{\uparrow}}
\newcommand{\rmse}{\text{RMSE} \lowerbetter}
\newcommand{\rmselog}{\text{RMSE}_{\log} \lowerbetter}
\newcommand{\fivepercent}{<5\% \higherbetter}
\newcommand{\tenpercent}{<10\% \higherbetter}
\newcommand{\fifteenpercent}{<15\% \higherbetter}
\newcommand{\absrel}{\text{Abs Rel} \lowerbetter}
\newcommand{\sqrel}{\text{Sq Rel} \lowerbetter}
\newcommand{\improves}[1]{{ \color{Highlight} {(#1)}}}
\newcommand{\Improves}[1]{{ \color{Highlight} (\textbf{#1})}}
\newcommand{\iMproves}[1]{{ \color{Highlight} (\underline{#1})}}

\newcommand{\tarref}{union}
\newcommand{\Tarref}{Union}
\newcommand{\geodist}{geo-distance}
\newcommand{\Geodist}{Geo-distance}

\title{R4D: Utilizing \underline{R}eference Objects \underline{for} Long-Range \underline{D}istance Estimation}

\author{Antiquus S.~Hippocampus, Natalia Cerebro \& Amelie P. Amygdale \thanks{ Use footnote for providing further information
about author (webpage, alternative address)---\emph{not} for acknowledging
funding agencies.  Funding acknowledgements go at the end of the paper.} \\
Department of Computer Science\\
Cranberry-Lemon University\\
Pittsburgh, PA 15213, USA \\
\texttt{\{hippo,brain,jen\}@cs.cranberry-lemon.edu} \\
\And
Ji Q. Ren \& Yevgeny LeNet \\
Department of Computational Neuroscience \\
University of the Witwatersrand \\
Joburg, South Africa \\
\texttt{\{robot,net\}@wits.ac.za} \\
\AND
Coauthor \\
Affiliation \\
Address \\
\texttt{email}
}

\begin{document}

\maketitle

%%%%%%%%% ABSTRACT
\begin{abstract}
Estimating the distance of objects is a safety-critical task for autonomous driving. 
% Under several challenging conditions, including autonomous trucking, freeway, or wet road driving, the autonomous vehicle needs to estimate the distance of multiple objects at large distances. 
Focusing on the short-range objects, existing methods and datasets neglect the equally important long-range objects. In this paper, we name this challenging but underexplored task as Long-Range Distance Estimation, and propose two datasets for this task.
% Existing state-of-the-art methods can accurately predict short-range distances, but are not well-suited for long-range objects beyond \lidar{} range.
We then propose \ourproposedmethod{}, the first framework to accurately estimate the distance of long-range objects by using references with known distances in the scene. Drawing inspiration from human perception, \ourproposedmethod{} builds a graph by connecting a target object to all references. An edge in the graph encodes the relative distance information between a pair of target and reference objects. An attention module is then used to weigh the importance of reference objects and combine them into one target object distance prediction. Experiments on the two proposed datasets demonstrate the effectiveness and robustness of \ourproposedmethod{} by showing significant improvements compared to existing baselines.
\end{abstract}

%%%%%%%%% BODY TEXT
\section{Introduction} \label{sec:intro}

% \begin{figure}[t]
% \centering
% \includegraphics[width=0.5\linewidth]{figs/LongRangeFigure1V5.pdf}
% % source: https://docs.google.com/drawings/d/1dLVlmSWTtZD8iA2Xp1d_5JEm8C0xA8zOdCuc6Ic4woY/edit?usp=sharing
% % Legal: Most vehicles (target objects) shown in this freeway view are beyond LiDAR range, but still important for SDC tasks.
% \caption{Most long-range \textcolor{TAR}{\textbf{vehicles}} (target objects) in this freeway view are beyond \textcolor{LIDAR}{\textbf{\lidar{}}} range, but are still important for safety. \Ourproposedmethod{} estimates the distance of \textcolor{TAR}{\textbf{target objects}} by using \textcolor{REF}{\textbf{reference objects}} with known distances (\eg, \lidar{} detections). \ywli{Adjust this figure.}} 
% \label{fig:teaser}
% \end{figure}

\begin{wrapfigure}{r}{0.37\textwidth}
  \begin{center}
    \vspace{-4ex}
    \includegraphics[width=0.98\linewidth]{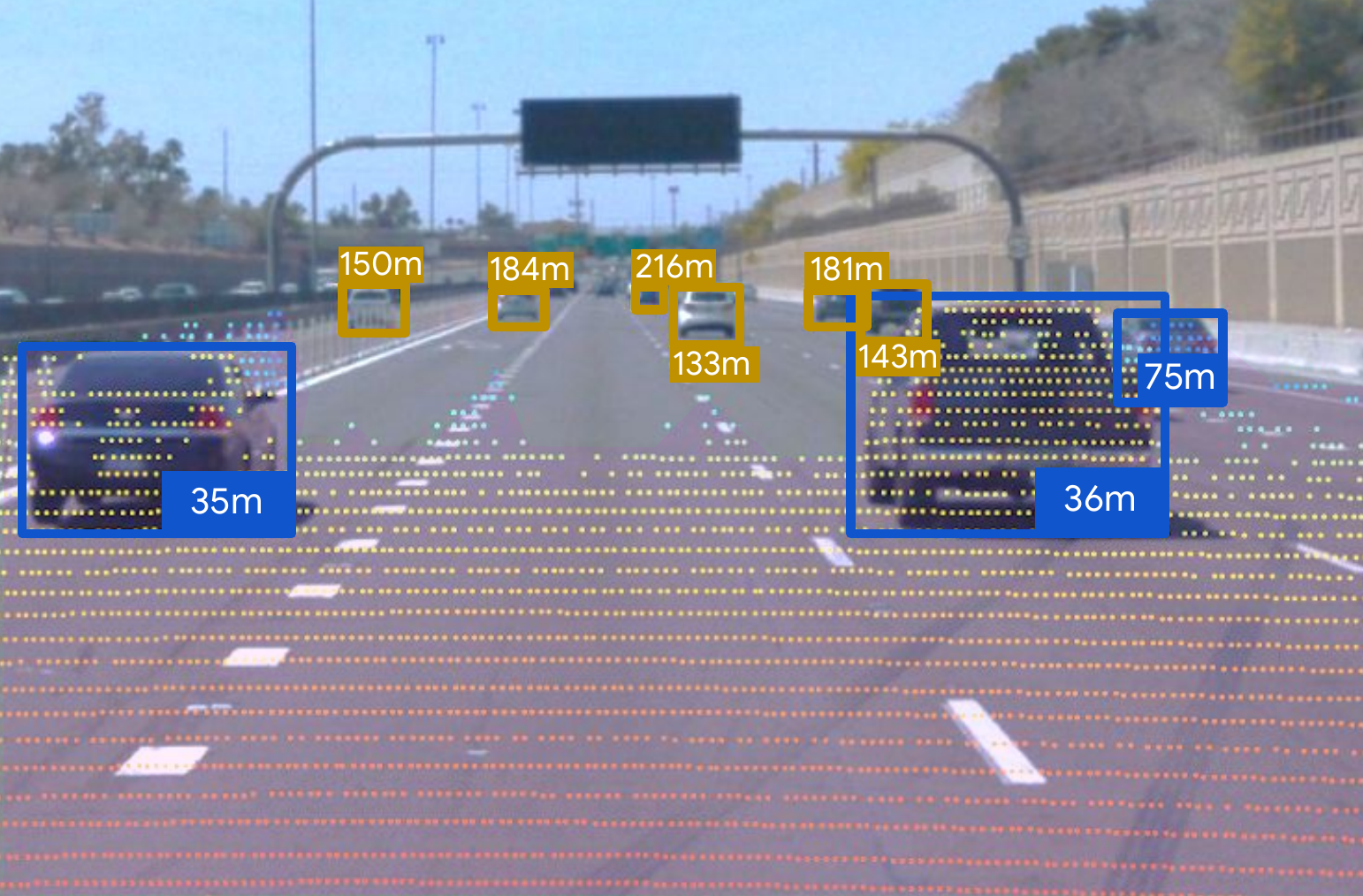}
    \vspace{-1ex}
    % source: https://docs.google.com/drawings/d/1dLVlmSWTtZD8iA2Xp1d_5JEm8C0xA8zOdCuc6Ic4woY/edit?usp=sharing
    % Legal: Most vehicles (target objects) shown in this freeway view are beyond LiDAR range, but still important for SDC tasks.
    \caption{Most long-range \textcolor{TAR}{\textbf{vehicles}} (target objects) in this  view are beyond \textcolor{LIDAR}{\textbf{\lidar{}}} range, but are still important for safety. \Ourproposedmethod{} estimates the distance of \textcolor{TAR}{\textbf{target objects}} by using \textcolor{REF}{\textbf{reference objects}} with known distances (\eg, \lidar{} detections).} 
    \label{fig:teaser}
    \vspace{-3ex}
  \end{center}
\end{wrapfigure}

% \begin{figure}[t]
% \centering
% \includegraphics[width=0.5\linewidth]{figs/LongRangeFigure1V5.pdf}
% % source: https://docs.google.com/drawings/d/1dLVlmSWTtZD8iA2Xp1d_5JEm8C0xA8zOdCuc6Ic4woY/edit?usp=sharing
% % Legal: Most vehicles (target objects) shown in this freeway view are beyond LiDAR range, but still important for SDC tasks.
% \caption{Most long-range \textcolor{TAR}{\textbf{vehicles}} (target objects) in this freeway view are beyond \textcolor{LIDAR}{\textbf{\lidar{}}} range, but are still important for safety. \Ourproposedmethod{} estimates the distance of \textcolor{TAR}{\textbf{target objects}} by using \textcolor{REF}{\textbf{reference objects}} with known distances (\eg, \lidar{} detections). \ywli{Adjust this figure.}} 
% \label{fig:teaser}
% \end{figure}
% \begin{figure}[b]
%   \begin{minipage}[c]{0.5\textwidth}
%     \includegraphics[width=1\linewidth]{figs/LongRangeFigure1V5.pdf}
%   \end{minipage}\hfill
%   \begin{minipage}[c]{0.47\textwidth}
%     \vspace{-1em}
%     \caption{Most long-range \textcolor{TAR}{\textbf{vehicles}} (target objects) in this freeway view are beyond \textcolor{LIDAR}{\textbf{\lidar{}}} range, but are still important for safety. \Ourproposedmethod{} estimates the distance of \textcolor{TAR}{\textbf{target objects}} by using \textcolor{REF}{\textbf{reference objects}} with known distances (\eg, \lidar{} detections).} \label{fig:teaser}
%   \end{minipage}
%   \vspace{-1em}
% \end{figure}

Estimating the distances of objects from the vehicle is crucial for several autonomous driving tasks, including lane changing, route planning, speed adjustment, collision avoidance, to name a few. Although existing methods and datasets focus on short-range objects, knowing the distance of long-range objects --- objects beyond a typical \lidar{} range of $\sim$80 meters (as shown in Figure~\ref{fig:teaser}) --- is necessary for freeway driving, heavy-duty truck driving, and wet road driving. Based on the US Department of Transportation~\citep{blanco2005visual}, on rural highways with a standard speed limit of 65 miles/h, it takes $\sim$145 meters for a passenger vehicle to come to a complete stop in an emergency, greatly exceeding the typical \lidar{} sensing range. The required stopping distance grows significantly with a heavy load truck or in bad road conditions such as snow, ice, or rain. For example, the stopping distance increases from 145 meters to 183 meters and 278 meters for trucking and wet road driving, respectively~\citep{longstoppingdistances,blanco2005visual}. 
 In addition, given that harsh and sudden breaking on freeways is unsafe, it remains critical to estimate the distance of objects beyond the minimum required stopping distance in order to provide enough time for a gradual slow-down or lane change. 
Therefore, to allow sufficient time for an appropriate reaction and to ensure safety, autonomous driving systems are required to estimate the distance to long-range objects. 
% In what follows, we will use the term \textit{localize} to refer to estimating the distance of an object with respect to the autonomous vehicle. 

\begin{figure*}[t]
\centering
\vspace{-2ex}
\includegraphics[width=0.73\linewidth]{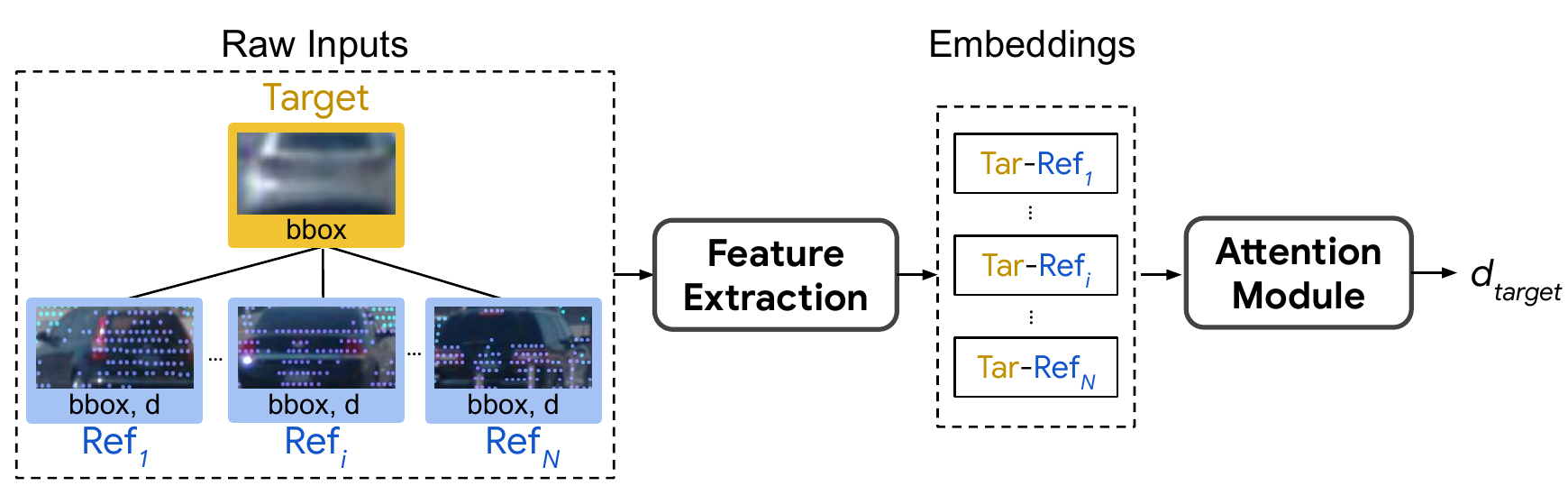}
% source: https://docs.google.com/drawings/d/1HE4mRrmJ2m6utJUgzMYRoeqteHpEsnlBxV_ttqjlk0k/edit?usp=sharing
% v4 source: https://docs.google.com/drawings/d/10E_tRUwGWvzpR43cbpjVMPixoP7LZl00cJL_0P9pmXQ/edit
\vspace{-2ex}
 \caption{\Ourproposedmethod{} overall architecture for training and prediction. Target and references are represented as a graph and the embeddings are fed to an attention module to weigh different references and aggregate results into a final distance estimate.}
\label{fig:abstract_arch}
\vspace{-4ex}
\end{figure*}

We name this critical but underexplored task as \textit{Long-Range Distance Estimation}. Concretely, given the short-range \lidar{} signal and the camera image, the distances of long-range objects (beyond \lidar{} range) are expected as output. Following conventions, the \textit{distance} is measured between the camera and the object center along the camera's optical axis~\citep{gokcce2015vision,haseeb2018disnet,zhu2019learning}. We discuss more about making design choices for our task in the related work. For this new task, we present \textit{pseudo long-range KITTI dataset} and \textit{\longrangdataset{} dataset}. Since KITTI~\citep{Geiger2013IJRR} do not provide ground-truth distance for long-range objects, the pseudo long-range KITTI dataset is a derived dataset by removing \lidar{} points beyond 40 meters and considering the objects beyond 40 meters as target. More importantly, we build a large scale dataset with annotated \textit{real} long-range objects (from 80 to 300 meters). In summary, with different definition of ``long-range" (either beyond 40 or 80 meters), both datasets include \lidar{} points, camera images, and distance labels for long-range objects.

Neither \lidar{} nor camera alone solves this long-range distance estimation problem to the desired accuracy. Most existing \textit{\lidar{}} technologies do not meet long-range sensing requirements. According to the Waymo and KITTI self-driving car datasets, the maximum \lidar{} range is only around 80 meters, which falls short of the range required for the aforementioned scenarios. Even though some advanced \lidar{} systems claim to achieve a longer sensing range,~\eg, Waymo's 5th generation \lidar{} system~\citep{waymo-ybr} and Velodyne Alpha Prime\texttrademark{}~\citep{alphaprime} which reach up to 300 meters, \lidar{} points are sparse at long-range and, thus, more likely to be occluded. 
Therefore, \lidar{} alone is not enough to cover all safety-critical self-driving car applications.

\textit{Cameras}, on the other hand, sense objects at a longer range and capture rich semantic information such as object appearance, geometry, and contextual hints. However, they do not, in and of themselves, provide depth information. Classical geometry-based algorithms could estimate the distance of canonical objects (\eg, sedans, trucks) based on their pixel sizes in the camera image~\citep{criminisi2000single,tuohy2010distance,gokcce2015vision,haseeb2018disnet,qi2019distance}. However, these approaches yield inaccurate results on long-range objects due to errors in size estimation.
Appearance-based methods~\citep{song2020endtoend,zhu2019learning} give unsatisfactory results on long-range objects. Such methods rely on a single appearance cue to estimate the distance, overlooking context or other relevant signals in the scene. At long-range, objects are visually small thus produce less informative appearance features. Although neither \lidar{} nor camera alone can solve this problem, these two signals provide complementary cues for long-range distance estimation.

In this paper, we propose \ourproposedmethod{}, a method to utilize \underline{r}eference objects with known and accurate distances \underline{for} long-range \underline{d}istance estimation (\ourproposedmethod{}). The main motivation behind our approach lies in theories of human perception from cognitive science: humans often estimate the distance of an object relative to other \textit{reference} points or objects~\citep{granrud1984comparison}. 
%Granrud~\etal study monocular depth perception in 5- and 7-month-old infants, and observe that cues about the size and distance of objects are determined relative to the size and distance of other objects~\citep{granrud1984comparison}. 
Specifically, we train a model to localize a long-range object (\textit{target}) given \textit{references} with known distances. We represent the target object and references as a \textit{graph}. As shown in Figure~\ref{fig:abstract_arch}, we define objects as nodes, with edges connecting the target to the reference objects. The reference information is propagated to the long-range target object by extracting target-reference (Tar-Ref) embeddings. \ourproposedmethod{} then feeds all target-reference embeddings to an attention module which fuses information from different references, by weighing their relative importance and combining them into one distance prediction.
% Existing autonomous driving datasets,~\eg,~KITTI~\citep{Geiger2013IJRR}, do not provide ground-truth distance for long-range objects. We build a pseudo long-range dataset based on KITTI by removing \lidar{} points beyond 40 meters. To further validate \ourproposedmethod{}, we build an \textit{\longrangdataset{}} dataset which contains 3D and 2D bounding boxes for long-range vehicles up to 300 meters. 

Experiments on the pseudo long-range KITTI and \longrangdataset{} datasets show that \ourproposedmethod{} significantly improves the distance estimation and achieves state-of-the-art long-range localization results. For example, on the \textit{\longrangdataset{}} dataset, with \ourproposedmethod{}, the distances of 8.9\% more vehicles (from 53.4\% to 62.3\%) are predicted with a relative distance error below 10\%. By conducting experiments on images captured at different times of the day,~\eg,~train on daytime and test on nighttime images, \ourproposedmethod{} also shows stronger robustness against domain changes.

To summarize, our contributions are three-fold. (1) We propose a critical but underexplored task Long-Range Distance Estimation. (2) We present two datasets, pseudo long-range KITTI dataset and \longrangdataset{} dataset. To facilitate future research, the datasets will be made publicly available.
% \footnote{The \longrangdataset{} dataset subjects to approvals.} 
(3) We develop \ourproposedmethod{}, the first framework to accurately estimate the distance of long-range objects by using references with known distances.

\section{Related Work}
\label{sec:related_work}

In this section, we discuss the related tasks and the detailed design choices of our proposed task. %We also discuss the attention mechanism, which is used in our proposed model \ourproposedmethod{}.
\paragraph{Monocular distance estimation.}
Estimating the distance of objects from an RGB image (\ie, monocular distance estimation) is a popular computer vision topic. More than a decade ago, researchers designed geometry-based algorithms to estimate object distances. For example, \citet{tuohy2010distance} use inverted perspective mapping (IPM) to convert an image to the bird's eye view for estimating distances. Learning-based methods were later proposed~\citep{qi2019distance,song2020endtoend}. For example, SVR~\citep{gokcce2015vision} and DisNet~\citep{haseeb2018disnet} take the pixel height and width of an object as input and estimate object distance by using support vector regression~\citep{drucker1997support} and multi-layer perceptron (MLP). However, such methods are subject to errors in object size estimation~\citep{zhu2019learning}, which can be even more pronounced for long-range objects. \citet{zhu2019learning} developed an end-to-end distance estimator with a Convolutional Neural Network (CNN) feature extractor. Specifically, given an RGB image and object bounding boxes, the model extracts per-object embeddings with a CNN and ROI pooling operator~\citep{girshick2015fast}, and then predicts a distance for each object. 
As previously mentioned, this method relies heavily on object appearance, which is a less informative cue for long-range objects.

Another related research topic is monocular per-pixel depth estimation. \citet{eigen2014depth,garg2016unsupervised,godard2019digging,lee2019monocular,liu2015learning,shu2020featdepth} propose to predict dense depth maps given RGB images. For example, \citet{lee2019monocular} generate multi-resolution relative depth maps to construct the final depth map. However, these methods can be resource-intensive and, therefore, difficult to incorporate within a latency-sensitive system such as autonomous driving~\citep{zhu2019learning}.
In addition, it is non-trivial to translate a depth map into per-object distance estimates due to occlusions, looseness of bounding boxes,~\etc. %It is also noteworthy to mention that several recent research findings illustrate the lack of robustness in standard dense depth map estimation methods~\cite{zhang2020adversarial,yamanaka2020adversarial,wong2020targeted,mathew2020monocular}.

\paragraph{Long-range depth sensing.}
In addition, other approaches such as stereo cameras and Radar have been used to sense objects at long-range. Stereo camera systems simulate human binocular vision and use multiple cameras to create stereo-pairs and perceive depth~\citep{khamis2018stereonet,poggi2019guided}. However, stereo vision systems have a few limitations: they are difficult to calibrate, are sensitive to vibrations~\citep{zhang2020depth}, and require a special hardware setup. 
In contrast, \ourproposedmethod{} does not need special hardware and works with existing sensors present on most autonomous vehicles, such as standalone camera and \lidar{}. %Besides, techniques used in \ourproposedmethod{} can be readily extended to other sensors and references.  
Radar, on the other hand, uses radio waves to determine the range, angle, and velocity of objects. However, due to its limited angular resolution~\citep{scheiner2020seeing}, it is difficult for Radar to accurately localize long-range objects.

\begin{figure*}[b]
\vspace{-3ex}
\centering
\includegraphics[width=0.98\linewidth]{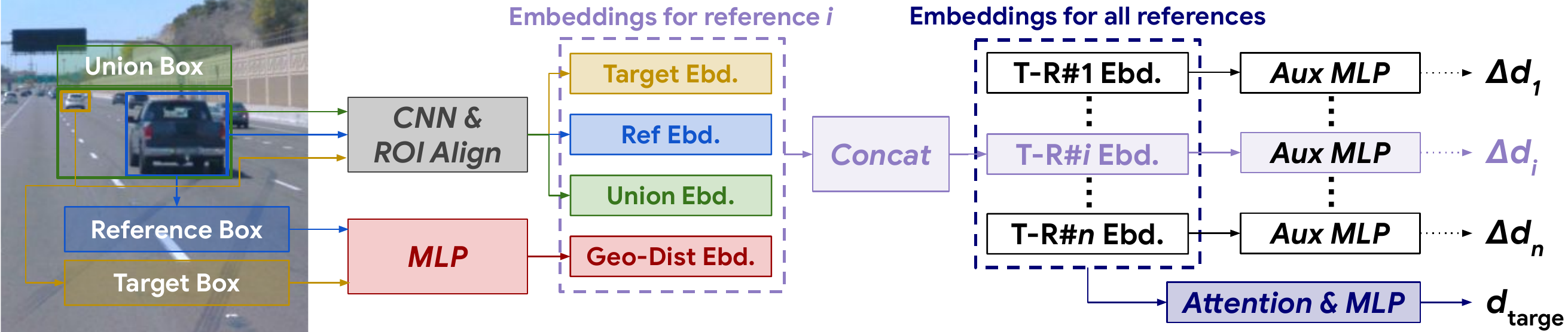}
% source: https://docs.google.com/drawings/d/1hG3QWGUDWrfGZlsL6CQgKpslGAEAc11mye78R0t54UI/edit?usp=sharing
% new source: https://docs.google.com/drawings/d/1zbjpWuEnS86fXerQIe8gVuiqtImPyKCnApKYuV2_TgU/edit?usp=sharing&resourcekey=0-Jb-Vq2aSdGvPBMlx9p4-qQ
% ICLR version: https://docs.google.com/drawings/d/17o_IwOOM1icrhR1vWyAVFTUPGCOx4-eHLnPGfx6X3uE/edit
\vspace{-1ex}
\caption{The detailed architecture for training and prediction. We use a CNN to extract a feature map from the input image. Given target, reference, and union bounding boxes, an ROIAlign is used to crop their corresponding embeddings from the feature map. Given the center and size of the target and reference boxes, we extract \geodist{} embeddings using a MLP. We refer to ``Attention \& MLP" and its output as the \textit{absolute distance head}, and ``Aux MLP" and its output as the \textit{relative distance head}. \textit{Relative heads} are only used during training. $\Delta d_i$ is the relative distance between the target and the $i^{\text{th}}$ reference object.} 
\label{fig:arch}
\vspace{-2ex}
\end{figure*}

\paragraph{Monocular 3D object detection.} 
% Monocular 3D object detection aims to detect all objects in an image by predicting a 3D bounding box for each of them,~\eg, MonoPair~\cite{chen2020monopair}. Besides distance (from the target object to the autonomous vehicle), the output also includes object size, orientation, amodal center (the projection of the object 3D center in the image plane). Take the object size for example. For short-range vehicles, these additional attributes like object size are critical for route planning and to avoid collisions. For long-range vehicles, the distance is significantly larger than object size (\eg,~ 300 meters vs. 1 meters), and the attributes such as size are less critical compared to the distance under the real-world scenario. However, these attributes increase the latency and the complexity of the model. \textit{Therefore, our task focuses on estimating the distance of long-range objects.}
Monocular 3D object detection aims to detect all objects in an image by predicting a 3D bounding box for each of them,~\eg, MonoPair~\citep{chen2020monopair}. Besides distance (from the target object to the autonomous vehicle), the monocular 3D object detector also predicts object size. For short-range vehicles, both object distance and object size are critical for self-driving tasks, such as route planning, collision avoidance, and lane changing. However, for long-range vehicles, the distance is significantly larger than object size (\eg,~300 meters vs. 2 meters), thus the object size is less critical compared to the distance under the real-world scenario. 
Moreover, predicting the object size increases the latency and the complexity of the model. \textit{Therefore, our task focuses on estimating the distance of long-range objects.}

\section{Methodology} 
\begin{wrapfigure}{r}{0.34\textwidth}
\vspace{-10ex}
\centering
\includegraphics[width=\linewidth]{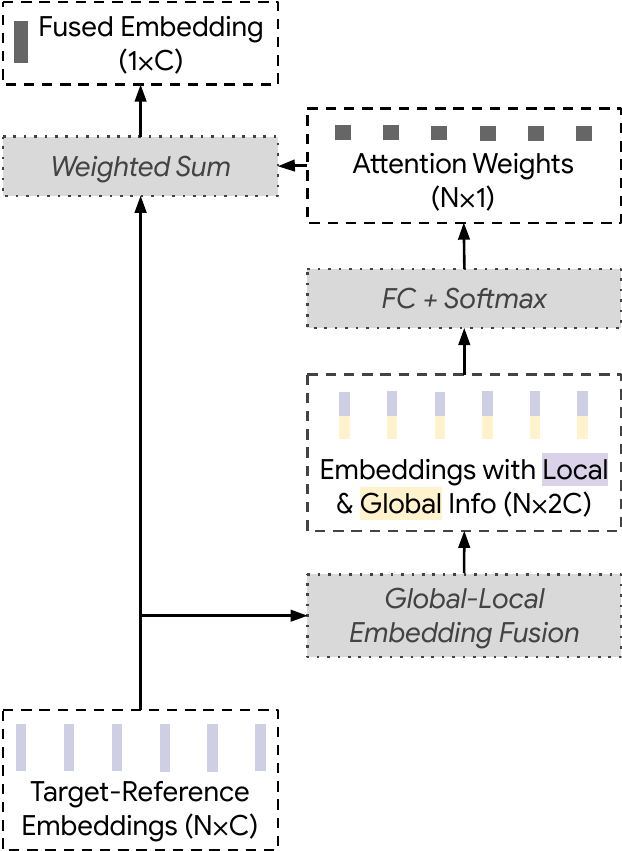}
\vspace{-3ex}
% source:  https://docs.google.com/drawings/d/1E6sqS3sVPNVUgFmxbpSiV9tRtAmbAPiIg0h3XKi0HAA/edit?usp=sharing
% ICLR version: https://docs.google.com/drawings/d/10d-KM0DfXWOkF2gwVTQ5j7PVme9UnkqiaHEtl9WtgWA/edit?usp=sharing&resourcekey=0-n3vf41r00AaonHGKBswO2A
\caption{Attention module to fuse target-reference embeddings.} 
\label{fig:attention}
\vspace{-3ex}
\end{wrapfigure}

\label{sec:methodology}
Motivated by the observation that humans estimate the distance of an object relative to other references~\citep{granrud1984comparison}, we propose \ourproposedmethod{}, a method utilizing \underline{R}eference objects \underline{For} long-range \underline{D}istance estimation. References can be any combination of objects or points with known and accurate distances to the autonomous vehicle, such as \lidar{} detections, objects detected by other sensors, map features,~\etc. We represent the target object and references as a \textit{graph} shown in the ``Raw Inputs" dashed box in Figure~\ref{fig:abstract_arch}. The target object and its references are nodes in this graph. Between a pair of target and reference objects, an edge is built to encode their pairwise relationship.  

The detailed architecture is illustrated in Figure~\ref{fig:arch}. To model the target-reference pairwise relationship, we propose to extract \textit{\tarref{} embeddings} and \textit{\geodist{} embeddings} which encode visual and geometry relationships, respectively (Section~\ref{sec:method_pair}).
Then, inspired by the intuition that reference objects are not equally important, we introduce an attention module in Section~\ref{sec:method_attention} to selectively aggregate pairwise relationships. 
Finally, %and inspired by residual representations~\cite{briggs2000multigrid,he2016deep,szeliski1990fast,szeliski2006locally},
\ourproposedmethod{} is trained with an auxiliary supervision: the \textit{relative distances} between the target and its references, as explained in  Section~\ref{sec:method_supervision}. 

In our setup, and without loss of generality, we use a monocular camera as the main sensor for detecting long-range target objects, and adopt \lidar{} detected short-range objects as references. It is worthy noting that \ourproposedmethod{} is not specifically designed for \lidar{} or monocular images and is readily extended to other sensors and references.

\subsection{Modeling Pairwise Relationships} \label{sec:method_pair}
As previously mentioned, to better estimate the distance of a target object, we propose to model the pairwise relationships between it and other reference objects.
For every pair of reference-target objects, we extract per-object feature embeddings and pairwise embeddings, as shown in Figure~\ref{fig:arch}. In particular, pairwise embeddings
include \textit{\tarref{}} embeddings and \textit{\geodist{}} embeddings which provide visual and geometry cues, respectively.

\paragraph{\Tarref{} embeddings.}
Visual cues for modeling pairwise relationships should be based on the target object, the reference object, as well as the scene between the two objects. We therefore propose a simple approach to extract feature embeddings from the union bounding box which covers both objects. Figure~\ref{fig:arch} shows an example of a union bounding box highlighted in green. This is similar to \citet{yao2018exploring}, where union bounding boxes were used to extract features for image captioning.

\paragraph{\Geodist{} embeddings.}
To provide geometry cues, we form an input using the following:\vspace{-0.2cm}
\begin{enumerate}[-]
    \item 2D bounding box \textit{center coordinates} of target and reference objects, and the relative position shift between them.\vspace{-0.2cm}
    \item 2D bounding box \textit{size} of target and reference objects, and the relative scale between them.\vspace{-0.2cm}
    \item The \textit{distance of the reference object} as provided by a \lidar{} 3D object detector.
\end{enumerate}
\vspace{-0.2cm}We, then, feed this input into a multi-layer perceptron to generate the \geodist{} embedding.
\; \newline

Given the target, reference, \tarref{}, and \geodist{} embeddings, we concatenate them to form a final target-reference embedding which models the relationship between a pair of target and reference objects. We analyze and discuss the importance of these embeddings in Section~\ref{sec:results_embeddings}.

\subsection{Attention-Based Information Aggregation} \label{sec:method_attention}
% \begin{figure}[t]
% \centering
% \includegraphics[width=0.5\linewidth]{figs/LongRangeFigureAttentionV2.pdf}
% % source:  https://docs.google.com/drawings/d/1E6sqS3sVPNVUgFmxbpSiV9tRtAmbAPiIg0h3XKi0HAA/edit?usp=sharing
% \caption{Attention module to fuse target-reference embeddings. \ywli{Adjust this figure.}} 
% \label{fig:attention}
% \end{figure}

% \begin{wrapfigure}{r}{0.44\textwidth}
%   \begin{center}
%     \vspace{-3ex}
%     \includegraphics[width=0.99\linewidth]{figs/LongRangeFigure1V5.pdf}
%     % source: https://docs.google.com/drawings/d/1dLVlmSWTtZD8iA2Xp1d_5JEm8C0xA8zOdCuc6Ic4woY/edit?usp=sharing
%     % Legal: Most vehicles (target objects) shown in this freeway view are beyond LiDAR range, but still important for SDC tasks.
%     \caption{Most long-range \textcolor{TAR}{\textbf{vehicles}} (target objects) in this freeway view are beyond \textcolor{LIDAR}{\textbf{\lidar{}}} range, but are still important for safety. \Ourproposedmethod{} estimates the distance of \textcolor{TAR}{\textbf{target objects}} by using \textcolor{REF}{\textbf{reference objects}} with known distances (\eg, \lidar{} detections).} 
%     \label{fig:teaser}
%     \vspace{-3ex}
%   \end{center}
% \end{wrapfigure}

Pairs of target-reference embeddings should be combined to estimate the distance of a given target object. A simple approach is to average them over all references of the same target object. However, intuitively, and as supported by our experimental results, reference objects are not equally important. For example, a car in the bottom left of the image is potentially less helpful when localizing a faraway car lying in the top right corner of the image.

In order to guide our model to focus on the most important reference objects, we introduce an attention-based module. As shown in Figure~\ref{fig:attention}, we follow VectorNet~\citep{gao2020vectornet} to construct embeddings with local and global information. Specifically, we use MLP and average pooling to extract a global embedding (shown in yellow in Figure~\ref{fig:attention}). The global embedding is, then, concatenated with the original target-reference embeddings (blue in Figure~\ref{fig:attention}). Given these global-local embeddings, the attention module predicts and normalizes importance weights for each reference using a fully connected layer followed by a softmax layer. It, finally, uses these weights to fuse the original target-reference embeddings, in a weighted average manner, resulting in one final embedding.

\subsection{Relative Distance Supervision} \label{sec:method_supervision}
Predicting the target distance from the target embedding is more straightforward than from other indirect cues such as reference embeddings. To encourage the model to learn the pairwise relationships between target and reference objects, instead of shortcut cues~\citep{geirhos2020shortcut,li2020shapetexture,wang2020makes}, %and inspired by Gradient Blending~\cite{wang2020makes}, 
we provide additional supervision. The design of this additional supervision is akin to the residual representation, which is widely used for computer vision to help the optimization procedure~\citep{briggs2000multigrid,he2016deep,szeliski1990fast,szeliski2006locally}. %In particular, the relative distance between a target and reference object serves as the \textit{residual} value between the target distance and the reference distance, as the additional supervision for each target-reference embedding.
Specifically, during the training stage, we add a relative (or residual) distance head for each target-reference embedding.  The \textit{relative} distance $\Delta d$ between the target object and the reference object is given by $\Delta d = d_t - d_r$, where $d_t$ (resp., $d_r$) is the distance between the camera and the target (resp., reference) object. 

\section{Datasets and Evaluation Metrics}
To evaluate on the new task, long-range distance estimation, we create \textit{pseudo long-range KITTI dataset} and \textit{\longrangdataset{} dataset}. We then introduce the evaluation metrics. %The details are elaborated below.

\subsection{Pseudo Long-Range KITTI Dataset} The original KITTI dataset~\citep{Geiger2013IJRR} contains RGB images, \lidar{} point clouds, as well as 2D and 3D bounding boxes and other per-pixel annotations. With these rich annotations, benchmarks for self-driving tasks have been well developed based on KITTI dataset, including scene flow estimation, depth estimation, detection, and segmentation.

\begin{figure}[tb]
  \begin{minipage}[tb]{0.33\textwidth}
  \centering
    \includegraphics[width=0.99\linewidth]{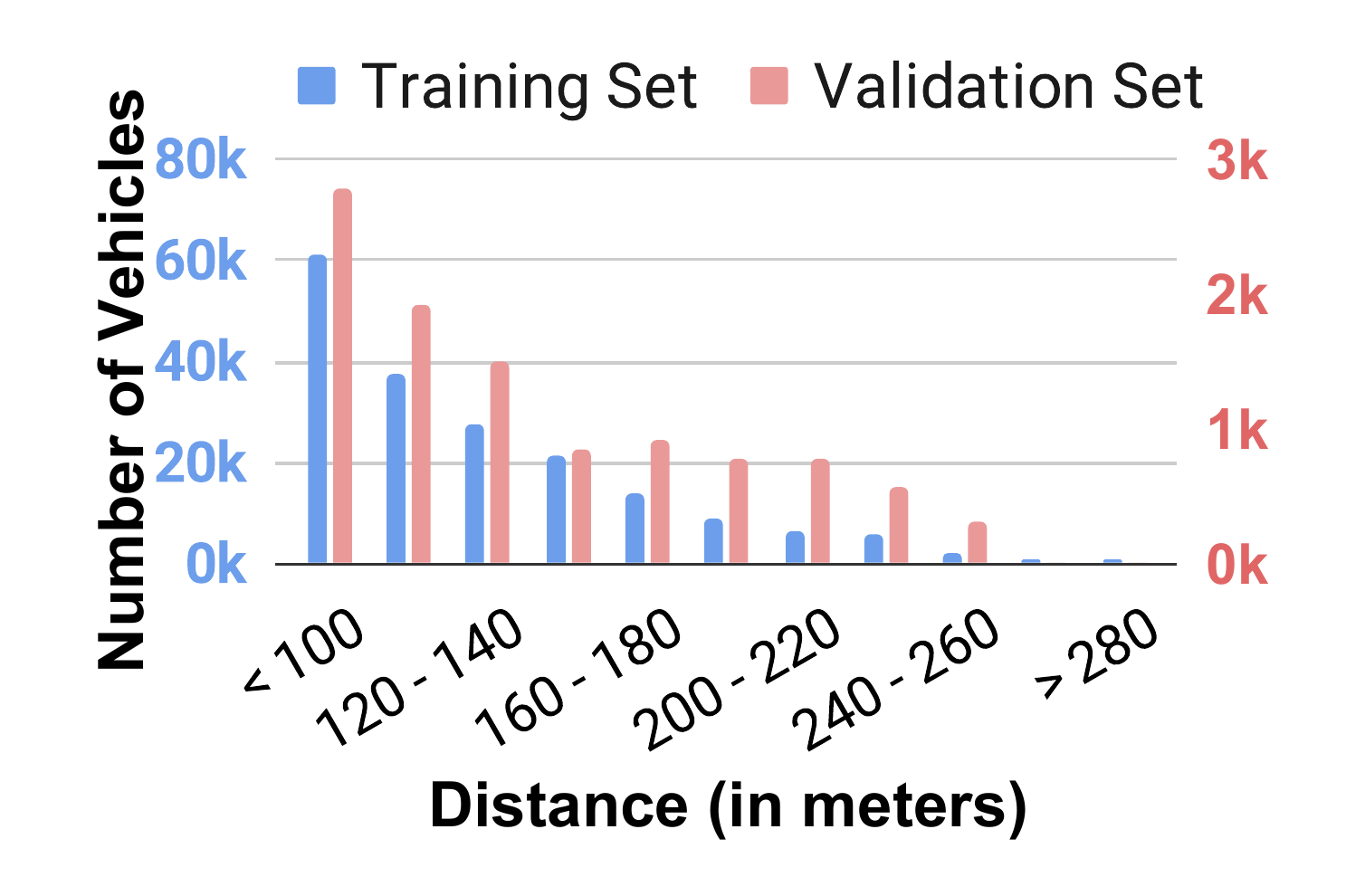}
    % \vspace{-5ex}
    \caption{Data distribution of the anonymous long-range dataset.} 
    \label{fig:dataset_stat}
  \end{minipage}
  \hfill
  \begin{minipage}[tb]{0.63\textwidth}
  \centering
    \includegraphics[width=0.99\linewidth]{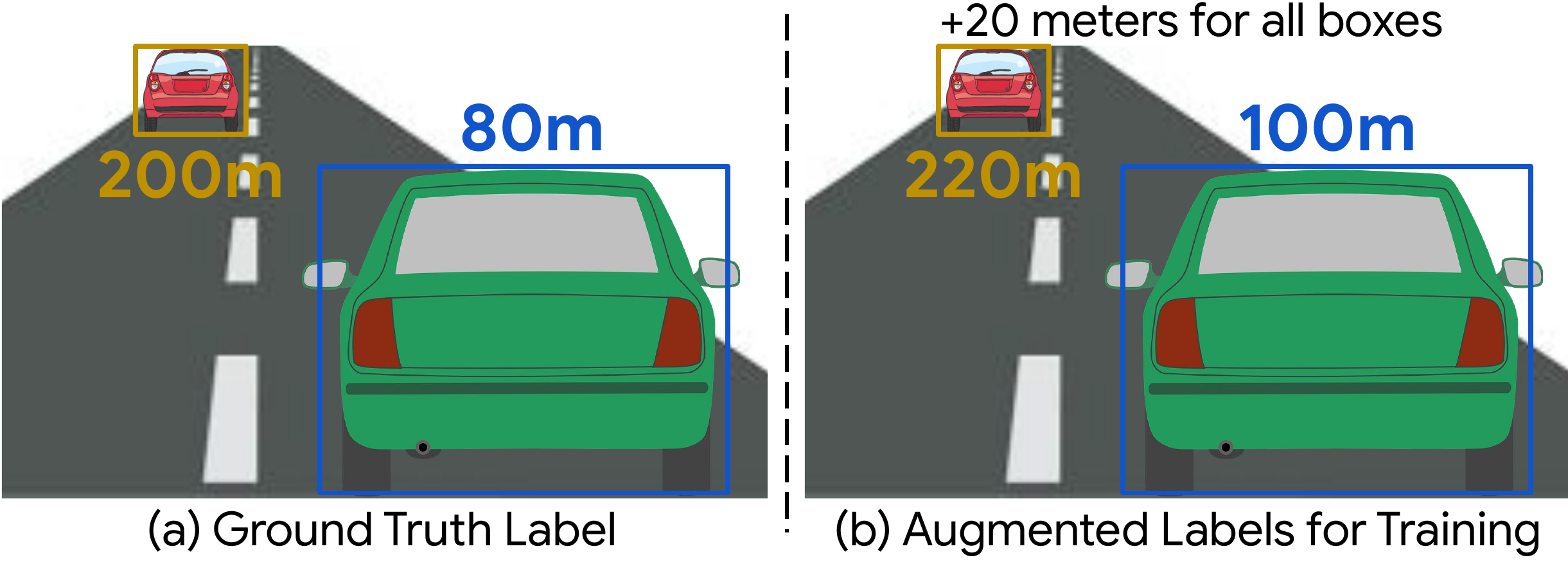}
    % \vspace{-3ex}
    \caption{An example of our distance augmentation for encouraging learning pairwise relationships.}
    \label{fig:dataaug}
  \end{minipage}
% \vspace{-3ex}
\end{figure}

Even though the KITTI dataset provides \lidar{} point clouds, it does not provide distance annotations beyond \lidar{} range. We hereby present \textit{pseudo long-range KITTI dataset}, which is derived from the original KITTI dataset. By assuming \lidar{} has an effective sensing range of 40 meters only, we remove \lidar{} points beyond it. The objects lying beyond this effective sensing range are considered as long-range target objects, while the distance of other objects are known and provided as inputs. It is worth noting that there are no overlap between the aforementioned target objects and the objects with known distance. Following the convention~\citep{shi2019pointrcnn}, the original training data is split into two subsets for training and validation, respectively. All the images do not contain long-range objects are removed from the dataset. Finally, the pseudo long-range KITTI dataset contains 2,181 images with 4,233 vehicles and 2,340 images with 4,033 vehicles in the training and validation set, respectively. Although it is easily obtained, this derived dataset is relatively small-scale and more critically does not annotate real long-range objects.

% \label{sec:exp}
% \begin{figure}[t]
% \centering
% \includegraphics[width=0.5\linewidth]{figs/dataset_stat.pdf}
% % https://docs.google.com/spreadsheets/d/1DK9NC1cdANksJORKFi406TrXMrg7wEhW2zRPrJ7GWqI/edit?usp=sharing
% \caption{The distribution of target vehicle distances in the Train and Val sets of the \longrangdataset{} dataset. \ywli{Adjust this figure.}} 
% \label{fig:dataset_stat}
% \end{figure}

% \begin{wrapfigure}{r}{0.43\textwidth}
%   \begin{center}
%     \vspace{-3ex}
%     \includegraphics[width=0.99\linewidth]{figs/dataset_stat.pdf}
%     % https://docs.google.com/spreadsheets/d/1DK9NC1cdANksJORKFi406TrXMrg7wEhW2zRPrJ7GWqI/edit?usp=sharing
%     \caption{The distribution of target vehicle distances in the Train and Val sets of the \longrangdataset{} dataset.} 
%     \label{fig:dataset_stat}
%     \vspace{-2ex}
%   \end{center}
% \end{wrapfigure}

% \begin{wrapfigure}{r}{0.55\textwidth}
%   \begin{center}
%     % \vspace{-15ex}
%     \includegraphics[width=1\linewidth]{figs/LongRangeFigureDataAugV3.pdf}
%     % source: https://docs.google.com/drawings/d/18ZjDuIUiu2ySSPcO_SlYMhmYqEIGzOUh8iOdFsYTbqo/edit?resourcekey=0-5JQlCor-tAIEImhkFISF4g
%     \caption{An example of our distance augmentation for encouraging learning pairwise relationships.}
%     \label{fig:dataaug}
%     % \vspace{-2ex}
%   \end{center}
% \end{wrapfigure}

\subsection{\LongRangDataset{} Dataset} Considering the aforementioned two limitations of the pseudo long-range KITTI dataset, we build a long-range dataset which includes vehicle distances up to 300 meters from the autonomous vehicle. The labels are obtained as follows. First, we create 3D boxes for objects in the \lidar{} range. Then, we use Radar derived signals for rough positioning and extend \lidar{} boxes to long-range. Due to the low resolution of Radar, labelers use smooth object trajectory constraints to make sure the boxes are consistent over time and space. Finally, given the 3D boxes for long-range objects, we compute their distances to the autonomous vehicle.
In total, as shown in Figure~\ref{fig:dataset_stat}, we obtain %353 training segments with 
49,056 training images with 187,938 long-range vehicles, and %32 validation segments with 
3,578 validation images containing 10,483 long-range vehicles. These images span  different times of day, including dawn, day, dusk, and night time. Considering its large scale and the real long-range annotations, we use this dataset by default.

\subsection{Evaluation Metrics} We follow the evaluation metrics used in dense depth and per-object distance estimation~\citep{eigen2014depth,liu2015learning,garg2016unsupervised,zhu2019learning,shu2020featdepth,zhang2020depth}, namely: 1) the percentage of objects with relative distance error below a certain threshold ($\fivepercent$, $\tenpercent$, $\fifteenpercent$), 2) the absolute relative difference ($\absrel$), 3) the squared relative difference ($\sqrel$), 4) the root of the mean squared error ($\rmse$), and 5) the root of the mean squared error computed from the logs of the predicted and ground-truth distances ($\rmselog$). Specifically, given ground-truth distances $d^*$ and predicted distances $d$, we compute the metrics as follows:
\begin{equation}
\begin{split}
{\absrel} = \mathrm{Average}\left(|d - d^*|/d^*\right),\quad \quad & ~~~~{\sqrel}  = \mathrm{Average}\left((d - d^*)^2/d^*\right), \\
~{\rmse}  = \sqrt{\mathrm{Average}\left((d - d^*)^2\right)}, \quad \quad & {\rmselog} = \sqrt{\mathrm{Average}\left((\log d - \log d^*)^2\right)}.
\end{split}
\end{equation}
Note that $\higherbetter$ and $\lowerbetter$ indicate whether a larger or smaller value of the corresponding metric indicates a better performance. We only evaluate performance on long-range objects.

\begin{table*}[t]
% \vspace{-2ex}
\renewcommand\arraystretch{0.8}
\centering
\footnotesize
\setlength\tabcolsep{4.4pt}
\begin{tabular}{l|c|ccc|cccc}
\toprule
Method & \lidar{} & $\fivepercent$ & $\tenpercent$ & $\fifteenpercent$ & $\absrel$ & $\sqrel$ & $\rmse$ & $\rmselog$ \\
\midrule
% IPM~\cite{tuohy2010distance} & \xmark & - & - & - & - & - & - & -  \\
% SVR~\citep{gokcce2015vision} & \xmark & 29.9\% & 51.6\% & 65.4\% & 12.8\% & 5.43 & 31.4 & 0.210 \\
SVR & \xmark & 29.9\% & 51.6\% & 65.4\% & 12.8\% & 5.43 & 31.4 & 0.210 \\
% DisNet~\citep{haseeb2018disnet} & \xmark & 31.3\% & 52.4\% & 66.5\% & 12.5\% & 5.28 & 31.3 & 0.207  \\
DisNet & \xmark & 31.3\% & 52.4\% & 66.5\% & 12.5\% & 5.28 & 31.3 & 0.207  \\
\citet{zhu2019learning} & \xmark & 31.2\% & 53.4\% & 68.7\% & 11.5\% & 4.43 & 29.4 & 0.187  \\
\citet{zhu2019learning} & \cmark & 34.1\% & 56.4\% & 70.5\% & 11.3\% & 4.78 & 31.3 & 0.194  \\
\Ourproposedmethod{}-NA (\textbf{ours}) & \cmark & \underline{35.9\%} & \underline{61.7\%} & \underline{76.4\%} & \underline{10.1\%} & \underline{3.68} & \underline{27.0} & \underline{0.168}  \\
\Ourproposedmethod{} (\textbf{ours}) & \cmark & \textbf{38.5\%} & \textbf{62.3\%} & \textbf{77.1\%} & \textbf{~~9.8\%} & \textbf{3.60} & \textbf{26.7} & \textbf{0.167} \\
\bottomrule
\end{tabular}
\vspace{-1ex}
\caption{Results on the \longrangdataset{} dataset. Compared to \citet{zhu2019learning}, \Ourproposedmethod{} achieves performance gains on all evaluation metrics. \Ourproposedmethod{} also outperforms SVR~\citep{gokcce2015vision}, and DisNet~\citep{haseeb2018disnet}. \Ourproposedmethod{}-NA refers to No-Attention \ourproposedmethod{}.}
\label{tab:main_results}
\vspace{-3ex}
\end{table*}

% \begin{figure}[t]
% \centering
% \includegraphics[width=1\linewidth]{figs/LongRangeFigureDataAugv2.pdf}
% % source: https://docs.google.com/drawings/d/1QrNH1gmzjo7C6Slng92GXVHw2hsOZFflrVL80jglY9o/edit?usp=sharing
% \caption{An example of our distance augmentation strategy for encouraging the model to learn pairwise relationships. \ywli{Clean this table}}
% \label{fig:dataaug}
% \end{figure}

\section{Experiments}

\subsection{Implementation Details} \label{sec:imp_details}
We put the detailed design choices in Section~\ref{sec:supp_imp_details}, and here we only introduce a critical component, distance augmentation, during the training stage.
% We use ResNet-50~\cite{he2016deep} followed by an ROIAlign layer~\cite{he2017mask} to extract target, reference, and \tarref{} embeddings. We then apply a fully connected layer to each embedding to reduce the feature dimension to 1,024. We use a three-layer perceptron to extract geo-distance embeddings. Each layer contains a single fully connected layer followed by layer normalization~\cite{ba2016layer} and ReLU~\cite{nair2010rectified}. For the attention module, we use the two-layer subgraph propagation operation mentioned in VectorNet~\cite{gao2020vectornet} as the Global-Local Embedding Fusion depicted in Figure~\ref{fig:attention}, and replace the max pooling layer with average pooling. The reference objects are detected using a multi-view fusion 3D object detector~\cite{zhou2020end} %trained on the Waymo Open Dataset
% with 72.1 3D AP (L1). When there is no reference object in an image (0.6\% of the images), we disable the relative distance supervision. We use smooth L1~\cite{huber1992robust} as the loss function~\cite{zhu2019learning}.
% We train \ourproposedmethod{} on an 8-core TPU with a total batch size of 32 images for 24 epochs. The learning rate is initially set to 0.0005 and then decays by a factor of 10 at the 16th and 22nd epochs. We warm up~\cite{goyal2017accurate} the learning rate for 1,800 iterations. We initialize the weights from a ResNet-50 model pretrained on ImageNet~\cite{he2016deep}. For the pseudo long-range KITTI dataset, we train over 150 epochs with an initial learning rate of 0.005 and cosine learning rate decay~\cite{loshchilov2016sgdr}.

\begin{table*}[b]
\vspace{-3ex}
\renewcommand\arraystretch{0.8}
\centering
\footnotesize
\setlength\tabcolsep{4.4pt}
\begin{tabular}{l|c|ccc|cccc}
\toprule
Method & \lidar{} & $ \fivepercent$ & $ \tenpercent$ & $ \fifteenpercent$ & $\absrel$ &$\sqrel$ & $\rmse$ & $\rmselog$\\
\midrule
% IPM~\cite{tuohy2010distance} & \xmark & - & - & - & - & - & - & -  \\
SVR & \xmark & 39.1\% & 65.9\% & \underline{79.0\%} & 10.2\% & 1.38 & ~~9.5 & 0.166  \\
DisNet & \xmark & 37.1\% & 65.0\% & 77.7\% & 10.6\% & 1.55 & 10.4 & 0.181  \\
\citet{zhu2019learning} & \xmark & 39.4\% & 65.8\% & 80.2\% & ~~\underline{8.7\%} & \underline{0.88} & ~~\underline{7.7} & \underline{0.131}  \\
\citet{zhu2019learning} & \cmark & \underline{41.1\%} & \underline{66.5\%} & 78.0\% & ~~8.9\% & 0.97 & ~~8.1 & 0.136  \\
\Ourproposedmethod{} (\textbf{ours}) & \cmark & \textbf{46.3\%} & \textbf{72.5\%} & \textbf{83.9\%} & \textbf{~~7.5\%} & \textbf{0.68} & \textbf{~~6.8} & \textbf{0.112}  \\
\bottomrule
\end{tabular}
\vspace{-3ex}
\caption{Results on the pseudo long-range KITTI dataset. \Ourproposedmethod{} outperforms SVR~\citep{gokcce2015vision}, DisNet~\citep{haseeb2018disnet}, and \citet{zhu2019learning}.}
\label{tab:kitti}
\end{table*}

\paragraph{Distance augmentation.}
Distance augmentation is designed for encouraging \ourproposedmethod{} to learn from pairwise relationships and prevent it from exclusively using a single cue (namely, the target embeddings).
%, referred to as shortcut learning~\cite{geirhos2020shortcut}
We guide the model to focus more on the relative distance between the reference and target objects by emphasizing the correlation between the pairwise embeddings and the distance prediction. Specifically, we keep the relative distance fixed, perturb the reference distance, and expect the model to predict the target distance with the same perturbation. For example, as shown in Figure~\ref{fig:dataaug}~(a), the relative distance between the target and reference object is 120 meters and given a reference distance of 80 meters, the model should predict a target distance of 200 meters (= 80 meters + 120 meters). As shown in Figure~\ref{fig:dataaug}~(b), when the provided reference distance is perturbed (by, say, 20 meters), the model should predict 220 meters (= 100 meters + 120 meters) as the target distance, in order to maintain the correct relative distance of 120 meters. This helps the model not overfit to appearance cues provided by the target object and be more robust to slight changes in camera parameters or field-of-view. Similar augmentation techniques have been successfully used to discourage shortcut learning in other deep learning tasks~\citep{li2020shapetexture}.
During training, the distance label perturbation is sampled from a Gaussian distribution $X \sim \mathcal{N}(\mu,\,\sigma^{2})\,$ with $\mu = 0$ and $\sigma = 200$ (See Section~\ref{sec:supp_dist} for details). For the pseudo long-range KITTI dataset, we use $\sigma = 50$. %Note that although distance augmentation increases the learning difficulty by involving noisy training labels, the well-trained models still predict unbiased distances because we set $\mu = 0$.

\begin{table*}[t]
\renewcommand\arraystretch{0.8}
\centering
% \footnotesize
\setlength\tabcolsep{4.4pt}
\begin{tabular}{cccc|ccc|cccc}
\toprule
 Tgt & Ref & Uni & G-D & $\fivepercent$ & $\tenpercent$ & $\fifteenpercent$ & $\absrel$ & $\sqrel$ & $\rmse$ & $\rmselog$\\
\midrule
\cmark & \xmark & \xmark & \xmark & 31.2\% & 53.4\% & 68.7\% & 11.5\% & 4.43 & 29.4 & 0.187  \\
\cmark & \cmark & \xmark & \xmark & 31.5\% & 60.8\% & 75.3\% & 11.2\% & 4.30 & 29.2 & 0.183 \\
\cmark & \cmark & \cmark & \xmark & 32.5\% & 55.2\% & 69.7\% & 11.2\% & 4.42 & 29.7 & 0.188 \\
\cmark & \cmark & \xmark & \cmark & 36.9\% & 60.8\% & 75.3\% & 10.2\% & 3.88 & 27.8 & 0.175 \\
\cmark & \cmark & \cmark & \cmark & \textbf{38.5\%} & \textbf{62.3\%} & \textbf{77.1\%} & \textbf{~~9.8\%} & \textbf{3.60} & \textbf{26.7} & \textbf{0.167} \\

\bottomrule
\end{tabular}
\vspace{-2ex}
\caption{The target embeddings (Tgt) and the reference embeddings (Ref) alone are not enough for modeling pairwise relationships. The union embeddings (Uni) and geo-distance embeddings (G-D) are also necessary components of \ourproposedmethod{}.}
\label{tab:abl_embeddings}
\vspace{-3ex}
\end{table*}

\begin{wrapfigure}{r}{0.5\textwidth}
  \begin{center}
    \vspace{-9.2ex}
    \includegraphics[width=0.98\linewidth]{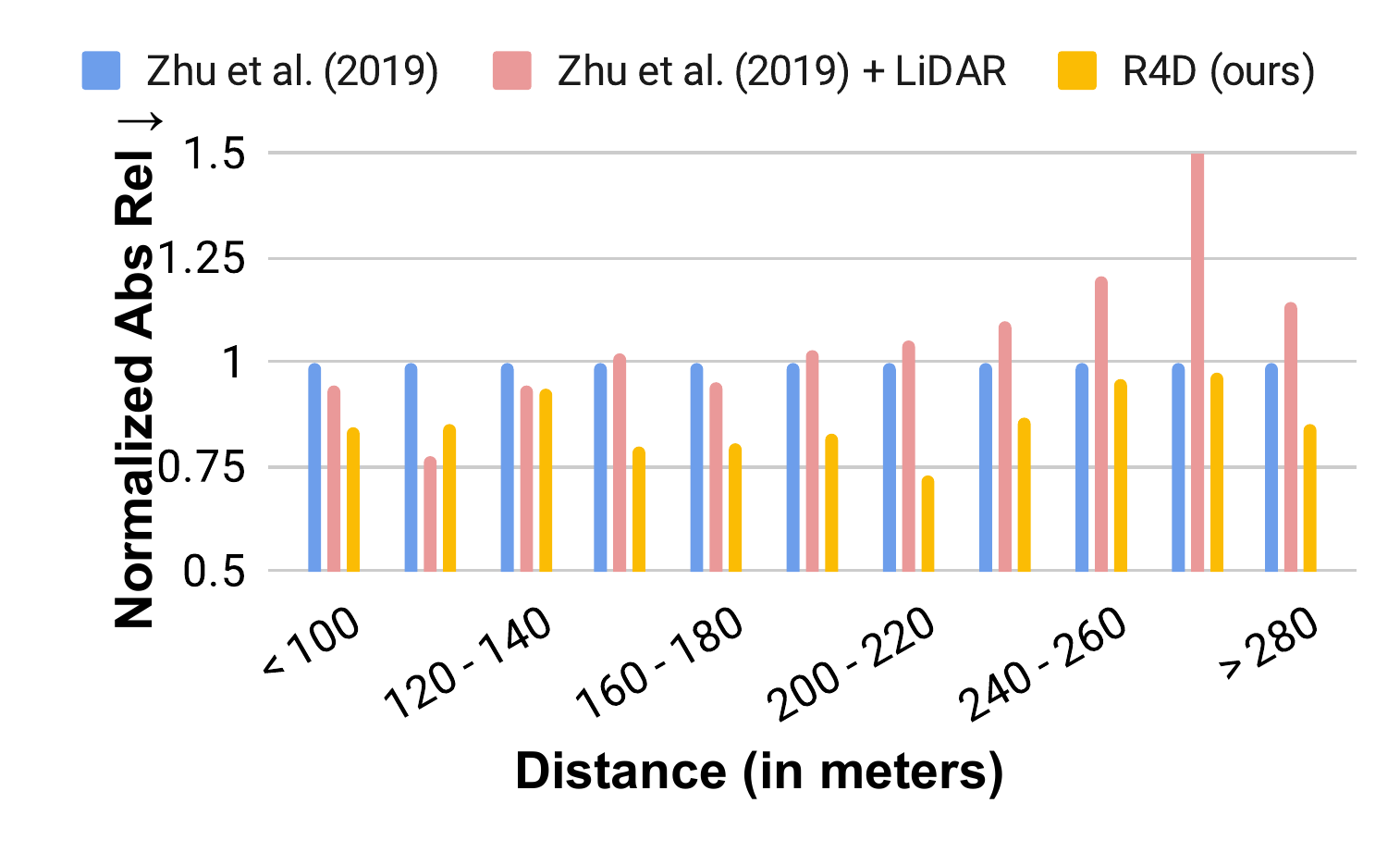}
    \vspace{-2ex}
    \caption{Comparison between \cite{zhu2019learning} with and without \lidar{} information. We show the $\absrel$ metric (normalized by the blue bars) for different ground-truth distance ranges. Results show that the simpler way of fusing short-range \lidar{} signals marginally improves the performance on objects slightly beyond \lidar{} range but hurts very long-range objects.} 
    \label{fig:abl_per_range_lidar}
    \vspace{-4ex}
  \end{center}
\end{wrapfigure}

\subsection{Long-Range Distance Estimation} \label{exp:main}
\label{sec:results}
Tables~\ref{tab:main_results} and~\ref{tab:kitti} show the distance estimation results on the \longrangdataset{} and pseudo long-range KITTI datasets. On both datasets, \ourproposedmethod{} significantly outperforms prior monocular distance estimation solutions. For example, compared to \citet{zhu2019learning}, \ourproposedmethod{} significantly improves the performance on the \longrangdataset{} dataset by 7.3\%, 8.9\%, and 8.4\% on the $\fivepercent$, $\tenpercent$, and $\fifteenpercent$ metrics, respectively. We observe similar trends on the pseudo long-range KITTI dataset: these improve by 6.9\%, 6.7\%, and 3.7\%, respectively. The improvement is even larger when compared to geometry-based techniques such as SVR~\citep{gokcce2015vision} and DisNet~\citep{haseeb2018disnet}. These improvements illustrate the effectiveness of \ourproposedmethod{}.

% \begin{figure}[t]
% \centering
% \includegraphics[width=1\linewidth]{figs/per_bucket.pdf}
% \caption{Comparison between \cite{zhu2019learning} with and without \lidar{} information. We show the $\absrel$ metric (normalized by the blue bars) for different ground-truth distance ranges. Results show that the simpler way of fusing short-range \lidar{} signals marginally improves the performance on objects slightly beyond \lidar{} range but hurts very long-range objects.} 
% \label{fig:teaser}
% \end{figure}

% \begin{figure}[tb]
% % \vspace{-3ex}
%   \begin{minipage}[c]{0.57\textwidth}
%     \includegraphics[width=1\linewidth]{figs/per_bucket.pdf}
%   \end{minipage}\hfill
%   \begin{minipage}[c]{0.37\textwidth}
%     \vspace{-4ex}
%     \caption{Comparison between \citet{zhu2019learning} with and without \lidar{} information. We show the $\absrel$ metric (normalized by the blue bars) for different ground-truth distance ranges. Results show that the simpler way of fusing short-range \lidar{} signals marginally improves the performance on objects slightly beyond \lidar{} range but hurts very long-range objects.}  \label{fig:abl_per_range_lidar}
%   \end{minipage}
% %   \vspace{-1em}
% \end{figure}

To further illustrate that \ourproposedmethod{} effectively uses references in \lidar{} range for long-range distance estimation, we design another simple baseline to use \lidar{} as a reference. We augment the input camera image (RGB) with a channel containing per-pixel projected \lidar{} distance values (D) (shown in Figure~\ref{fig:teaser}). Then, given an RGB+D image, and detected bounding boxes, the model predicts a distance value for each target object. We modify the input of our baseline, \citet{zhu2019learning}, to include the distance channel (D), in addition to the RGB image. However, as the results shown in Tables~\ref{tab:main_results}~and~\ref{tab:kitti} (row 4), this approach achieves similar performance as when the camera image (RGB) is the sole input. 
% This suggests that this model is unable to extrapolate distance values beyond the \lidar{} range. 
Besides, in Figure~\ref{fig:abl_per_range_lidar}, we show the distance estimation results of \ourproposedmethod{} as well as \citet{zhu2019learning} with RGB and RGB+D inputs, broken-down over different ground-truth distance ranges. We note that the addition of the \lidar{} distance channel only improves the performance of objects slightly beyond \lidar{} range, but fails at extrapolating distance values for localizing extremely long-range objects. In contrast, \ourproposedmethod{} offers consistent performance gains compared to \citet{zhu2019learning} across all distances. This results demonstrates the importance of explicitly modeling object relationships (the strategy of \ourproposedmethod{}), compared to the simpler baseline which has access to the same \lidar{} sensor data.

\begin{table*}[t]
\vspace{-2ex}
\footnotesize
\setlength\tabcolsep{1.8pt}
\renewcommand\arraystretch{0.8}
\centering
\begin{tabular}{l|cccc|ccc|cccc}
\toprule
Method & Abs. & Rel.  & \#Ref & DA & $\fivepercent$ & $\tenpercent$ & $\fifteenpercent$ & $\absrel$ & $\sqrel$ & $\rmse$ & $\rmselog$\\
\midrule
\citet{zhu2019learning} & \cmark & \xmark & Zero & \xmark & 32.9\% & 55.6\% & 69.9\% & 11.2\% & 4.36 & 29.3 & 0.187  \\
\citet{zhu2019learning} & \cmark & \xmark & Zero & \cmark & 32.3\% & 54.3\% & 69.4\% & 11.4\% & 4.37 & 29.2 & 0.186  \\ \midrule
\Ourproposedmethod{}-Relative  & \xmark & \cmark & One & \xmark & 34.4\% & 58.8\% & 73.8\% & 10.6\% & 4.00 & 28.1 & 0.178 \\
\Ourproposedmethod{}-Relative  & \xmark & \cmark & One & \cmark & \underline{34.9\%} & 57.4\% & 73.3\% & 10.6\% & 3.93 & 27.8 & \underline{0.174} \\ \midrule
\Ourproposedmethod{}-One-Ref  & \cmark & \cmark & One & \xmark & \textbf{37.5\%} & \underline{59.9\%} & \underline{74.3\%} & 10.4\% & 3.92 & 27.7 & 0.176 \\
\Ourproposedmethod{}-One-Ref  & \cmark & \cmark & One & \cmark & \textbf{37.5\%} & 59.5\% & 73.8\% & \underline{10.3\%} & \underline{3.87} & \underline{27.6} & 0.175 \\ \midrule
\Ourproposedmethod{}  & \cmark & \cmark & All & \xmark & 32.0\% & 54.1\% & 70.9\% & 11.1\% & 4.30 & 29.3 & 0.187  \\
\Ourproposedmethod{}  & \cmark & \cmark & All & \cmark & \textbf{37.5\%} & \textbf{63.3\%} & \textbf{78.6\%} & \textbf{~~9.8\%} & \textbf{3.64} & \textbf{26.9} & \textbf{0.169} \\

\bottomrule
\end{tabular}
% \vspace{-2ex}
\caption{Ablation studies on the proposed pairwise relationship modeling and distance augmentation (DA). Results in this table are slightly different from those in Tables~\ref{tab:main_results}~and~\ref{tab:abl_embeddings} because, here, we remove images which do not contain any reference objects. This is in order to report results on \ourproposedmethod{}-Relative, which requires at least one reference object (see Section~\ref{sec:exp_ref} for details). }
\label{tab:abl_dataaug}\label{tab:abl_rel_vs_abs}
% \vspace{-1ex}
\end{table*}

\begin{figure}[b]
% \vspace{-2ex}
\includegraphics[width=1\linewidth]{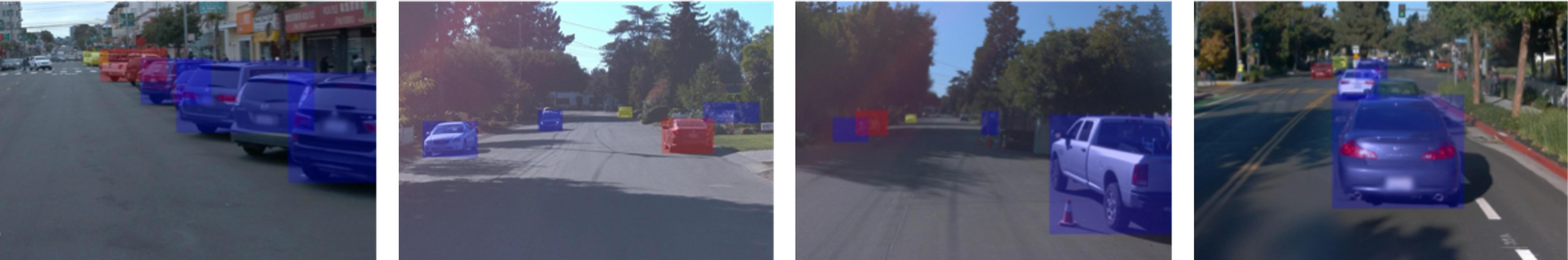}
% \vspace{-4ex}
\caption{Visualizing the learnt reference weights. Targets are highlighted in yellow. References with the largest attention weights are highlighted in red. Other references are colored in blue. We observe that the most important references are often close to or in the same lane as the target, and share a similar pose as the target. Best viewed zoomed in. More images are shown in Section~\ref{sec:vis_attention}.}
\label{fig:vis_attention}
\end{figure}

\subsection{Ablation Studies}
\paragraph{Attention-based information aggregation.} 
As discussed in Section~\ref{sec:method_attention}, given a fixed target, \ourproposedmethod{} aggregates all target-reference embeddings using an attention-based module and predicts a weight for each reference. As shown in Table~\ref{tab:main_results}, compared to treating all references equally and taking the average of all target-reference embeddings (R4D-NA), the attention module improves the performance across all metrics, \eg, $<$5\%\textcolor{red}{$\uparrow$} is improved by 2.6\%.

In order to inspect the attention module output, we visualize the learnt reference weights. We highlight the reference object with the largest weight in Figure~\ref{fig:vis_attention}. We find that results match our intuition. For example, the most important references usually share the same lane and pose as the target. In addition, the best reference is usually the one closest to the target object.

\paragraph{Efficacy of feature embeddings.} \label{sec:results_embeddings}
In this section, we perform ablation experiments on the proposed feature embeddings (Section~\ref{sec:method_pair}) to show their importance. The first two rows in Table~\ref{tab:abl_embeddings} show that, when only using the target and reference embeddings, the performance is sub-optimal. This is likely due to the fact that these embeddings fail to capture pairwise relationships and only encode individual appearance, pose, and geometry information. Incorporating the \geodist{} embeddings (row 4) allows the model to learn from the relative position and geometry between target and reference objects, resulting in the largest performance improvement. Finally, adding the \tarref{} embeddings (row 5), which encode relative appearance cues, further improves the performance (\eg,~$\fivepercent$ improves by 1.6\%). Note that the \tarref{} embeddings are only helpful when used together with the \geodist{} embeddings. This is because the union bounding box may include multiple potential objects, and a precise position is critical to pinpoint the reference object location.

\paragraph{Benefits of modeling pairwise relationships.} \label{sec:exp_ref}
\ourproposedmethod{}'s main performance improvements stem from modeling pairwise relationships between target and reference objects. In this section, we show how incorporating references helps. We study the effect of removing the relative distance supervision (\ie, the ``Aux MLP'' head in Figure~\ref{fig:arch}) or the absolute distance supervision. When only the relative distance head is enabled, we use its output during inference and the target distance is computed by adding the relative distance to the known reference distance (we exclude images without references). 
%We, therefore, exclude images where no references are present.
% .

As shown in Table~\ref{tab:abl_rel_vs_abs}, randomly choosing one reference (row 3) and predicting the distance using relative distance achieves a $\tenpercent$ of 58.8\%, outperforming the baseline~\citep{zhu2019learning} which directly predicts the absolute distance (row 1 with a $\tenpercent$ of 55.6\%). %Since the absolute and relative heads capture complementary information,
Combining the absolute and relative heads during training (row 5) further improves the $\tenpercent$ metric by 4.3\% (achieving a $\tenpercent$ metric of 59.9\%). Finally, this improvement is the largest when multiple references are aggregated using the attention module (row 8), resulting in a $\tenpercent$ of 63.3\%. 

\paragraph{Distance augmentation.}
As described in Section~\ref{sec:imp_details}, we propose the distance augmentation strategy to encourage \ourproposedmethod{} to learn better pairwise relationships during training.
Table~\ref{tab:abl_dataaug} shows the results from ablation experiments related to distance augmentation. The distance augmentation is only critical for our final model \ourproposedmethod{}, which (1) includes both absolute and relative distance prediction heads, and (2) considers \textit{multiple} reference objects. Without distance augmentation (row 7), the model does not learn pairwise relationships, and the performance of \ourproposedmethod{}'s absolute distance head drops to the baseline's performance (row 1). 
In fact, when at most one reference is used during training (\citet{zhu2019learning}, \ourproposedmethod{}-Relative, and \ourproposedmethod{}-One-Ref), the proposed distance augmentation marginally affects the performance. Our intuition is that the one- or zero-reference cases require less supervision and data diversity due to the simpler input and model structures.

\begin{table*}[t]
\renewcommand\arraystretch{0.8}
\centering
\footnotesize
\setlength\tabcolsep{3pt}
\begin{tabular}{l|l|HHlllll}
\toprule
Valid. Set & Method & $\fivepercent$ & $\tenpercent$ & $\fifteenpercent$ & $\absrel$ & $\sqrel$ & $\rmse$ & $\rmselog$ \\
\midrule
\multirow{2}{*}{Day} & \citet{zhu2019learning} & 31.6\% & 54.7\% & 72.2\% & 11.2\% & 4.52 & 30.3 & 0.190 \\
 & \Ourproposedmethod{} (\textbf{ours}) & 38.1\%\Improves{+6.5\%} & 62.9\%\improves{+8.2\%} & 77.0\%\improves{+4.8\%} & ~~9.6\%\improves{-1.6\%} & 3.71\improves{-0.81} & 27.8\improves{-2.5} & 0.171\improves{-0.019} \\
\midrule
Dawn \& & \citet{zhu2019learning} & 27.7\% & 48.6\% & 63.8\% & 13.6\% & 6.48 & 36.4 & 0.233 \\
Dusk & \Ourproposedmethod{} (\textbf{ours}) & 32.2\%\improves{+4.5\%} & 55.9\%\improves{+7.3\%} & 70.0\%\iMproves{+6.2\%} & 11.3\%\iMproves{-2.3\%} & 4.19\iMproves{-2.29} & 28.5\iMproves{-7.9} & 0.177\iMproves{-0.056} \\
\midrule
\multirow{2}{*}{Night} & \citet{zhu2019learning} & 14.0\% & 25.6\% & 36.8\% & 23.3\% & 12.5 & 46.8 & 0.344 \\
 & \Ourproposedmethod{} (\textbf{ours}) & 17.8\%\improves{+3.8\%} & 34.8\%\Improves{+9.2\%} & 49.1\%\Improves{+12.3\%} & 17.7\%\Improves{-5.6\%} & ~~8.2\Improves{-4.38} & 38.4\Improves{-6.5} & 0.267\Improves{-0.077} \\
\bottomrule
\end{tabular}
% \vspace{-1ex}
\caption{Robustness on out-of-distribution data. Both \ourproposedmethod{} and the baseline~\citep{zhu2019learning} are trained on daytime images only, and evaluated on three test sets collected during different times of day. The performance improvement of \ourproposedmethod{} increases when the domain gap is larger.}
\label{tab:robustness}
% \vspace{-2ex}
\end{table*}

\subsection{Robustness on Out-Of-Distribution Data}
For safety-critical applications, such as autonomous driving, the ability to handle out-of-distribution (OOD) data is crucial~\citep{vyas2018out}. We expect \ourproposedmethod{} to be more robust on OOD data because it captures contextual information, in addition to appearance and geometry cues. In contrast, baseline methods~\citep{zhu2019learning} use  a single prediction cue, thus rendering them more sensitive to appearance changes. To validate the robustness on OOD data, we train our model solely on images captured during daytime, but evaluate on three different times of day, namely: daytime, dawn/dusk, and night-time. Specifically, we split the \longrangdataset{} dataset into four disjoint sets: \textit{train-day} (used for training), and \textit{val-day}, \textit{val-dawn\&dusk}, \textit{val-night} (used for validation). 
The results of this experiment are shown in Table~\ref{tab:robustness}. Compared to the baseline method~\citep{zhu2019learning}, \ourproposedmethod{} achieves a larger improvement when the domain gap between the training and validation sets is larger. For example, when evaluating on daytime (small domain gap) and night-time (large domain gap), \ourproposedmethod{} achieves an $\absrel$ metric improvement of 1.6\% and 5.6\%, respectively.

\section{Conclusion}
% Estimating the distance of objects at long-range is a crucial task for autonomous driving. 
In this paper, we present an underexplored task, long-range distance estimation, as well as two datasets for it. We then introduced \ourproposedmethod{}, a novel approach to address this problem. Our method can accurately localize long-range objects using references with known distances in the scene.
Experiments on two proposed datasets demonstrate that it significantly outperforms prior solutions and achieves state of the art long-range distance estimation performance. 
It should also be noted that \ourproposedmethod{} is not specifically designed for \lidar{} or monocular images and is readily extended to other sensors and references, such as features from high-definition maps. It can also be used to incorporate temporal information by fusing past object detections as references, which is a direction we hope to explore to further improve the long-range distance estimation performance.

\bibliography{iclr2022_conference}

\begin{thebibliography}{47}
\providecommand{\natexlab}[1]{#1}
\providecommand{\url}[1]{\texttt{#1}}
\expandafter\ifx\csname urlstyle\endcsname\relax
  \providecommand{\doi}[1]{doi: #1}\else
  \providecommand{\doi}{doi: \begingroup \urlstyle{rm}\Url}\fi

\bibitem[Administration(2016)]{longstoppingdistances}
Federal Motor Carrier~Safety Administration.
\newblock Long stopping distances, {Federal Motor Carrier Safety
  Administration}.
\newblock \url{https://www.fmcsa.dot.gov/ourroads/long-stopping-distances},
  2016.
\newblock Accessed: Sep-09-2021.

\bibitem[Ba et~al.(2016)Ba, Kiros, and Hinton]{ba2016layer}
Jimmy~Lei Ba, Jamie~Ryan Kiros, and Geoffrey~E Hinton.
\newblock Layer normalization.
\newblock \emph{arXiv preprint arXiv:1607.06450}, 2016.

\bibitem[Blanco \& Hankey(2005)Blanco and Hankey]{blanco2005visual}
Myra Blanco and Jonathan~M Hankey.
\newblock Visual performance during nighttime driving in fog.
\newblock Technical report, FHWA-HRT-04-137, 2005.

\bibitem[Briggs et~al.(2000)Briggs, Henson, and McCormick]{briggs2000multigrid}
William~L Briggs, Van~Emden Henson, and Steve~F McCormick.
\newblock \emph{A multigrid tutorial}.
\newblock SIAM, 2000.

\bibitem[Caesar et~al.(2019)Caesar, Bankiti, Lang, Vora, Liong, Xu, Krishnan,
  Pan, Baldan, and Beijbom]{nuscenes2019}
Holger Caesar, Varun Bankiti, Alex~H. Lang, Sourabh Vora, Venice~Erin Liong,
  Qiang Xu, Anush Krishnan, Yu~Pan, Giancarlo Baldan, and Oscar Beijbom.
\newblock {nuScenes}: A multimodal dataset for autonomous driving.
\newblock \emph{arXiv preprint arXiv:1903.11027}, 2019.

\bibitem[Chen et~al.(2020)Chen, Tai, Sun, and Li]{chen2020monopair}
Yongjian Chen, Lei Tai, Kai Sun, and Mingyang Li.
\newblock Monopair: Monocular {3D} object detection using pairwise spatial
  relationships.
\newblock In \emph{Conference on Computer Vision and Pattern Recognition
  (CVPR)}, 2020.

\bibitem[Criminisi et~al.(2000)Criminisi, Reid, and
  Zisserman]{criminisi2000single}
Antonio Criminisi, Ian Reid, and Andrew Zisserman.
\newblock Single view metrology.
\newblock \emph{International Journal of Computer Vision (IJCV)}, 40\penalty0
  (2):\penalty0 123--148, 2000.

\bibitem[Drucker et~al.(1997)Drucker, Burges, Kaufman, Smola, and
  Vapnik]{drucker1997support}
Harris Drucker, Christopher~JC Burges, Linda Kaufman, Alex~J Smola, and
  Vladimir Vapnik.
\newblock Support vector regression machines.
\newblock In \emph{Advances in Neural Information Processing Systems
  (NeurIPS)}, 1997.

\bibitem[Eigen et~al.(2014)Eigen, Puhrsch, and Fergus]{eigen2014depth}
David Eigen, Christian Puhrsch, and Rob Fergus.
\newblock Depth map prediction from a single image using a multi-scale deep
  network.
\newblock In \emph{Advances in Neural Information Processing Systems
  (NeurIPS)}, 2014.

\bibitem[Gao et~al.(2020)Gao, Sun, Zhao, Shen, Anguelov, Li, and
  Schmid]{gao2020vectornet}
Jiyang Gao, Chen Sun, Hang Zhao, Yi~Shen, Dragomir Anguelov, Congcong Li, and
  Cordelia Schmid.
\newblock {VectorNet}: Encoding {HD} maps and agent dynamics from vectorized
  representation.
\newblock In \emph{Conference on Computer Vision and Pattern Recognition
  (CVPR)}, 2020.

\bibitem[Garg et~al.(2016)Garg, Bg, Carneiro, and Reid]{garg2016unsupervised}
Ravi Garg, Vijay~Kumar Bg, Gustavo Carneiro, and Ian Reid.
\newblock Unsupervised {CNN} for single view depth estimation: Geometry to the
  rescue.
\newblock In \emph{European Conference on Computer Vision (ECCV)}, 2016.

\bibitem[Geiger et~al.(2013)Geiger, Lenz, Stiller, and Urtasun]{Geiger2013IJRR}
Andreas Geiger, Philip Lenz, Christoph Stiller, and Raquel Urtasun.
\newblock Vision meets robotics: The {KITTI} dataset.
\newblock \emph{International Journal of Robotics Research (IJRR)}, 2013.

\bibitem[Geirhos et~al.(2020)Geirhos, Jacobsen, Michaelis, Zemel, Brendel,
  Bethge, and Wichmann]{geirhos2020shortcut}
Robert Geirhos, J{\"o}rn-Henrik Jacobsen, Claudio Michaelis, Richard Zemel,
  Wieland Brendel, Matthias Bethge, and Felix~A Wichmann.
\newblock Shortcut learning in deep neural networks.
\newblock \emph{arXiv preprint arXiv:2004.07780}, 2020.

\bibitem[Girshick(2015)]{girshick2015fast}
Ross Girshick.
\newblock Fast {R-CNN}.
\newblock In \emph{International Conference on Computer Vision (ICCV)}, 2015.

\bibitem[Godard et~al.(2019)Godard, Mac~Aodha, Firman, and
  Brostow]{godard2019digging}
Cl{\'e}ment Godard, Oisin Mac~Aodha, Michael Firman, and Gabriel~J Brostow.
\newblock Digging into self-supervised monocular depth estimation.
\newblock In \emph{International Conference on Computer Vision (ICCV)}, 2019.

\bibitem[G{\"o}k{\c{c}}e et~al.(2015)G{\"o}k{\c{c}}e, {\"U}{\c{c}}oluk,
  {\c{S}}ahin, and Kalkan]{gokcce2015vision}
Fatih G{\"o}k{\c{c}}e, G{\"o}kt{\"u}rk {\"U}{\c{c}}oluk, Erol {\c{S}}ahin, and
  Sinan Kalkan.
\newblock Vision-based detection and distance estimation of micro unmanned
  aerial vehicles.
\newblock \emph{Sensors}, 15\penalty0 (9):\penalty0 23805--23846, 2015.

\bibitem[Goyal et~al.(2017)Goyal, Doll{\'a}r, Girshick, Noordhuis, Wesolowski,
  Kyrola, Tulloch, Jia, and He]{goyal2017accurate}
Priya Goyal, Piotr Doll{\'a}r, Ross Girshick, Pieter Noordhuis, Lukasz
  Wesolowski, Aapo Kyrola, Andrew Tulloch, Yangqing Jia, and Kaiming He.
\newblock Accurate, large minibatch {SGD}: Training {ImageNet} in 1 hour.
\newblock \emph{arXiv preprint arXiv:1706.02677}, 2017.

\bibitem[Granrud et~al.(1984)Granrud, Yonas, and
  Pettersen]{granrud1984comparison}
Carl~E Granrud, Albert Yonas, and Linda Pettersen.
\newblock A comparison of monocular and binocular depth perception in 5-and
  7-month-old infants.
\newblock \emph{Journal of Experimental Child Psychology}, 38\penalty0
  (1):\penalty0 19--32, 1984.

\bibitem[Haseeb et~al.(2018)Haseeb, Guan, Risti{\'c}-Durrant, and
  Gr{\"a}ser]{haseeb2018disnet}
Muhammad~Abdul Haseeb, Jianyu Guan, Danijela Risti{\'c}-Durrant, and Axel
  Gr{\"a}ser.
\newblock {DisNet}: A novel method for distance estimation from monocular
  camera.
\newblock \emph{10th Planning, Perception and Navigation for Intelligent
  Vehicles (PPNIV18), IROS}, 2018.

\bibitem[He et~al.(2016)He, Zhang, Ren, and Sun]{he2016deep}
Kaiming He, Xiangyu Zhang, Shaoqing Ren, and Jian Sun.
\newblock Deep residual learning for image recognition.
\newblock In \emph{Conference on Computer Vision and Pattern Recognition
  (CVPR)}, 2016.

\bibitem[He et~al.(2017)He, Gkioxari, Doll{\'a}r, and Girshick]{he2017mask}
Kaiming He, Georgia Gkioxari, Piotr Doll{\'a}r, and Ross Girshick.
\newblock Mask {R-CNN}.
\newblock In \emph{International Conference on Computer Vision (ICCV)}, 2017.

\bibitem[Huber(1992)]{huber1992robust}
Peter~J Huber.
\newblock Robust estimation of a location parameter.
\newblock In \emph{Breakthroughs in statistics}, pp.\  492--518. Springer,
  1992.

\bibitem[Jeyachandran(2020)]{waymo-ybr}
Satish Jeyachandran.
\newblock {Waymo's 5th Generation Self-Driving LiDAR}.
\newblock
  \url{https://blog.waymo.com/2020/03/introducing-5th-generation-waymo-driver.html},
  2020.
\newblock Accessed: Sep-09-2021.

\bibitem[Khamis et~al.(2018)Khamis, Fanello, Rhemann, Kowdle, Valentin, and
  Izadi]{khamis2018stereonet}
Sameh Khamis, Sean Fanello, Christoph Rhemann, Adarsh Kowdle, Julien Valentin,
  and Shahram Izadi.
\newblock Stereonet: Guided hierarchical refinement for real-time edge-aware
  depth prediction.
\newblock In \emph{European Conference on Computer Vision (ECCV)}, 2018.

\bibitem[Lee \& Kim(2019)Lee and Kim]{lee2019monocular}
Jae-Han Lee and Chang-Su Kim.
\newblock Monocular depth estimation using relative depth maps.
\newblock In \emph{Conference on Computer Vision and Pattern Recognition
  (CVPR)}, 2019.

\bibitem[Li et~al.(2020)Li, Zhao, Liu, and Cao]{li2020rtm3d}
Peixuan Li, Huaici Zhao, Pengfei Liu, and Feidao Cao.
\newblock Rtm3d: Real-time monocular 3d detection from object keypoints for
  autonomous driving.
\newblock In \emph{European Conference on Computer Vision (ECCV)}, 2020.

\bibitem[Li et~al.(2021)Li, Yu, Tan, Mei, Tang, Shen, Yuille, and
  Xie]{li2020shapetexture}
Yingwei Li, Qihang Yu, Mingxing Tan, Jieru Mei, Peng Tang, Wei Shen, Alan
  Yuille, and Cihang Xie.
\newblock Shape-texture debiased neural network training.
\newblock In \emph{International Conference on Learning Representations}, 2021.

\bibitem[Liu et~al.(2015)Liu, Shen, Lin, and Reid]{liu2015learning}
Fayao Liu, Chunhua Shen, Guosheng Lin, and Ian Reid.
\newblock Learning depth from single monocular images using deep convolutional
  neural fields.
\newblock \emph{IEEE Transactions on Pattern Analysis and Machine Intelligence
  (PAMI)}, 38\penalty0 (10):\penalty0 2024--2039, 2015.

\bibitem[Liu et~al.(2020)Liu, Wu, and T{\'o}th]{liu2020smoke}
Zechen Liu, Zizhang Wu, and Roland T{\'o}th.
\newblock {SMOKE}: Single-stage monocular {3D} object detection via keypoint
  estimation.
\newblock In \emph{Conference on Computer Vision and Pattern Recognition (CVPR)
  Workshops}, 2020.

\bibitem[Loshchilov \& Hutter(2017)Loshchilov and Hutter]{loshchilov2016sgdr}
Ilya Loshchilov and Frank Hutter.
\newblock Sgdr: Stochastic gradient descent with warm restarts.
\newblock In \emph{ICLR}, 2017.

\bibitem[Nair \& Hinton(2010)Nair and Hinton]{nair2010rectified}
Vinod Nair and Geoffrey~E Hinton.
\newblock Rectified linear units improve restricted {Boltzmann} machines.
\newblock In \emph{International Conference on Machine Learning (ICML)}, 2010.

\bibitem[Poggi et~al.(2019)Poggi, Pallotti, Tosi, and
  Mattoccia]{poggi2019guided}
Matteo Poggi, Davide Pallotti, Fabio Tosi, and Stefano Mattoccia.
\newblock Guided stereo matching.
\newblock In \emph{Conference on Computer Vision and Pattern Recognition
  (CVPR)}, 2019.

\bibitem[Qi et~al.(2019)Qi, Li, Sun, Zhang, and Sun]{qi2019distance}
SH~Qi, J~Li, ZP~Sun, JT~Zhang, and Y~Sun.
\newblock Distance estimation of monocular based on vehicle pose information.
\newblock In \emph{J. Phys., Conf. Ser.}, 2019.

\bibitem[Scheiner et~al.(2020)Scheiner, Kraus, Wei, Phan, Mannan, Appenrodt,
  Ritter, Dickmann, Dietmayer, Sick, et~al.]{scheiner2020seeing}
Nicolas Scheiner, Florian Kraus, Fangyin Wei, Buu Phan, Fahim Mannan, Nils
  Appenrodt, Werner Ritter, Jurgen Dickmann, Klaus Dietmayer, Bernhard Sick,
  et~al.
\newblock Seeing around street corners: Non-line-of-sight detection and
  tracking in-the-wild using doppler radar.
\newblock In \emph{Conference on Computer Vision and Pattern Recognition},
  2020.

\bibitem[Shi et~al.(2019)Shi, Wang, and Li]{shi2019pointrcnn}
Shaoshuai Shi, Xiaogang Wang, and Hongsheng Li.
\newblock Pointrcnn: 3d object proposal generation and detection from point
  cloud.
\newblock In \emph{Conference on Computer Vision and Pattern Recognition
  (CVPR)}, 2019.

\bibitem[Shu et~al.(2020)Shu, Yu, Duan, and Yang]{shu2020featdepth}
Chang Shu, Kun Yu, Zhixiang Duan, and Kuiyuan Yang.
\newblock Feature-metric loss for self-supervised learning of depth and
  egomotion.
\newblock In \emph{European Conference on Computer Vision (ECCV)}, 2020.

\bibitem[{Song} et~al.(2020){Song}, {Lu}, {Zhang}, and {Li}]{song2020endtoend}
Z.~{Song}, J.~{Lu}, T.~{Zhang}, and H.~{Li}.
\newblock End-to-end learning for inter-vehicle distance and relative velocity
  estimation in adas with a monocular camera.
\newblock In \emph{International Conference on Robotics and Automation (ICRA)},
  2020.

\bibitem[Szeliski(1990)]{szeliski1990fast}
Richard Szeliski.
\newblock Fast surface interpolation using hierarchical basis functions.
\newblock \emph{IEEE Transactions on Pattern Analysis and Machine Intelligence
  (PAMI)}, 12\penalty0 (6):\penalty0 513--528, 1990.

\bibitem[Szeliski(2006)]{szeliski2006locally}
Richard Szeliski.
\newblock Locally adapted hierarchical basis preconditioning.
\newblock \emph{ACM Transactions on Graphics (TOG)}, 25\penalty0 (3):\penalty0
  1135--1143, 2006.

\bibitem[Tuohy et~al.(2010)Tuohy, O'Cualain, Jones, and
  Glavin]{tuohy2010distance}
S~Tuohy, D~O'Cualain, E~Jones, and M~Glavin.
\newblock Distance determination for an automobile environment using inverse
  perspective mapping in opencv.
\newblock In \emph{IET Irish Signals and Systems Conference (ISSC)}, 2010.

\bibitem[VelodyneLidar(2021)]{alphaprime}
VelodyneLidar.
\newblock {Alpha Prime™}.
\newblock \url{https://velodynelidar.com/products/alpha-prime/}, 2021.
\newblock Accessed: Sep-09-2021.

\bibitem[Vyas et~al.(2018)Vyas, Jammalamadaka, Zhu, Das, Kaul, and
  Willke]{vyas2018out}
Apoorv Vyas, Nataraj Jammalamadaka, Xia Zhu, Dipankar Das, Bharat Kaul, and
  Theodore~L Willke.
\newblock Out-of-distribution detection using an ensemble of self supervised
  leave-out classifiers.
\newblock In \emph{European Conference on Computer Vision (ECCV)}, 2018.

\bibitem[Wang et~al.(2020)Wang, Tran, and Feiszli]{wang2020makes}
Weiyao Wang, Du~Tran, and Matt Feiszli.
\newblock What makes training multi-modal classification networks hard?
\newblock In \emph{Conference on Computer Vision and Pattern Recognition
  (CVPR)}, 2020.

\bibitem[Yao et~al.(2018)Yao, Pan, Li, and Mei]{yao2018exploring}
Ting Yao, Yingwei Pan, Yehao Li, and Tao Mei.
\newblock Exploring visual relationship for image captioning.
\newblock In \emph{European Conference on Computer Vision (ECCV)}, 2018.

\bibitem[Zhang et~al.(2020)Zhang, Xie, Snavely, and Chen]{zhang2020depth}
Kai Zhang, Jiaxin Xie, Noah Snavely, and Qifeng Chen.
\newblock Depth sensing beyond {LiDAR} range.
\newblock In \emph{Conference on Computer Vision and Pattern Recognition
  (CVPR)}, 2020.

\bibitem[Zhou et~al.(2020)Zhou, Sun, Zhang, Anguelov, Gao, Ouyang, Guo, Ngiam,
  and Vasudevan]{zhou2020end}
Yin Zhou, Pei Sun, Yu~Zhang, Dragomir Anguelov, Jiyang Gao, Tom Ouyang, James
  Guo, Jiquan Ngiam, and Vijay Vasudevan.
\newblock End-to-end multi-view fusion for {3D} object detection in {LiDAR}
  point clouds.
\newblock In \emph{Conference on Robot Learning}, 2020.

\bibitem[Zhu \& Fang(2019)Zhu and Fang]{zhu2019learning}
Jing Zhu and Yi~Fang.
\newblock Learning object-specific distance from a monocular image.
\newblock In \emph{International Conference on Computer Vision (ICCV)}, 2019.

\end{thebibliography}
\bibliographystyle{iclr2022_conference}

\appendix
% \section{Appendix}
\section{Implementation Details} \label{sec:supp_imp_details}
We use ResNet-50~\citep{he2016deep} followed by an ROIAlign layer~\citep{he2017mask} to extract target, reference, and \tarref{} embeddings. We then apply a fully connected layer to each embedding to reduce the feature dimension to 1,024. We use a three-layer perceptron to extract geo-distance embeddings. Each layer contains a single fully connected layer followed by layer normalization~\citep{ba2016layer} and ReLU~\citep{nair2010rectified}. For the attention module, we use the two-layer subgraph propagation operation mentioned in VectorNet~\citep{gao2020vectornet} as the Global-Local Embedding Fusion depicted in Figure~4 in the main submission, and replace the max pooling layer with average pooling. The reference objects are detected using a multi-view fusion 3D object detector~\citep{zhou2020end} %trained on the Waymo Open Dataset
with 72.1 3D AP (L1). When there is no reference object in an image (0.6\% of the images), we disable the relative distance supervision. We use smooth L1~\citep{huber1992robust} as the loss function~\citep{zhu2019learning}.
We train \ourproposedmethod{} on an 8-core TPU with a total batch size of 32 images for 24 epochs. The learning rate is initially set to 0.0005 and then decays by a factor of 10 at the 16th and 22nd epochs. We warm up~\citep{goyal2017accurate} the learning rate for 1,800 iterations. We initialize the weights from a ResNet-50 model pretrained on ImageNet~\citep{he2016deep}. For the pseudo long-range KITTI dataset, we train over 150 epochs with an initial learning rate of 0.005 and cosine learning rate decay~\citep{loshchilov2016sgdr}.

\section{Hyper-parameter Sensitivity: Distance Augmentation} \label{sec:supp_dist}
In this section, we show how the performance of \ourproposedmethod{} varies with the distance augmentation hyper-parameter $\sigma$. In our final model, we perturbed the ground-truth distances of target and reference objects by a Gaussian random variable with distribution $X \sim \mathcal{N}(\mu,\,\sigma^{2})\,$ with $\mu = 0$ and $\sigma = 200$. In Table~\ref{tab:sensibility}, we show the performance of our method for different values of $\sigma$ ranging from 1 to 300. We observe that the proposed distance augmentation consistently improves the performance of \ourproposedmethod{} for all values of $\sigma$. The model reaches its optimal performance when $\sigma$ is around 200.

\begin{table*}[t]
\renewcommand\arraystretch{0.8}
\centering
\begin{tabular}{c|c|ccc|cccc}
\toprule
DistAug & $\sigma$ & $\fivepercent$ & $\tenpercent$ & $\fifteenpercent$ & $\absrel$ & $\sqrel$ & $\rmse$ & $\rmselog$ \\
\midrule
\xmark & N/A & 32.0\% & 54.1\% & 70.9\% & 11.1\% & 4.30 & 29.3 & 0.187  \\
\cmark & 1  & 32.8\% & 56.5\% & 71.9\% & 11.0\% & 4.32 & 29.2 & 0.186 \\
\cmark & 2  & 32.6\% & 54.9\% & 71.6\% & 10.9\% & 4.13 & 28.7 & 0.182 \\
\cmark & 5  & 32.4\% & 55.0\% & 71.0\% & 11.2\% & 4.44 & 29.8 & 0.189 \\
\cmark & 10  & 33.9\% & 56.8\% & 73.3\% & 10.7\% & 4.18 & 29.0 & 0.183 \\
\cmark & 20  & 35.7\% & 58.1\% & 73.4\% & 10.3\% & 3.89 & 28.0 & 0.175 \\
\cmark & 50  & 36.6\% & 59.5\% & 75.1\% & 10.2\% & 3.91 & 28.1 & 0.175 \\
\cmark & 100 & 34.6\% & 58.3\% & 75.0\% & 10.4\% & 3.91 & 28.0 & 0.176  \\
\cmark & 150 & \underline{37.5\%} & \underline{62.1\%} & \underline{76.6\%} & ~~\underline{9.9\%} & \textbf{3.61} & \textbf{26.8} & \textbf{0.167}  \\
\cmark & 200 & \underline{37.5\%} & \textbf{63.3\%} & \textbf{78.6\%} & ~~\textbf{9.8\%} & \underline{3.64} & \underline{26.9} & \underline{0.169} \\
\cmark & 250 & \textbf{38.3\%} & 61.0\% & 75.2\% & 10.1\% & 3.79 & 27.5 & 0.173  \\
\cmark & 300 & 37.2\% & 60.7\% & 75.6\% & 10.1\% & 3.77 & 27.4 & 0.171  \\
\bottomrule
\end{tabular}
\caption{Hyper-parameter sensitivity of distance augmentation.}
\label{tab:sensibility}
\end{table*}

\section{Inference Latency and Number of References}
In the main draft, we set the maximum number of references to 50. In this section, we show the performance and inference time of our method when we vary the maximum number of references. We draw two conclusions from the results (in Table~\ref{tab:latency}). First, more references lead to better results. For example, we observe the with a maximum of 2 reference objects, the  $\fivepercent$ metric improves by 2\% (from 35.0\% to 37.0\%) compared to using a single reference object.
In addition, when the number of references is large (say, 5), the benefit of using additional reference objects is marginal. For example, we observe that the $\fivepercent$ metric increases by only 0.3\% (from 37.2\% to 37.5\%) when the maximum number of references is increased from 5 to 50.
 Finally, and as expected, the latency increases with the number of reference objects. However, this increase is not significant when using a relatively small number of references. For example, when we increase the maximum number of references from 1 to 5, the latency only increases by 0.6 ms, while the $\tenpercent$ metric improves by 9.2\%.

\begin{table*}[t]
\small
\renewcommand\arraystretch{0.8}
\setlength\tabcolsep{3pt}
\centering
\begin{tabular}{c|c|ccc|cccc|c}
\toprule
Method & \#Ref & $\fivepercent$ & $\tenpercent$ & $\fifteenpercent$ & $\absrel$ & $\sqrel$ & $\rmse$ & $\rmselog$ & T (ms)$\lowerbetter$ \\
\midrule
\cite{zhu2019learning} & 0 & 32.0\% & 54.1\% & 70.9\% & 11.1\% & 4.30 & 29.3 & 0.187 & \textbf{15.0}  \\
\ourproposedmethod (\textbf{ours}) & 1  & 35.0\% & 60.7\% & 76.0\% & 10.4\% & 3.93 & 27.9 & 0.175 & \underline{15.2} \\
\ourproposedmethod (\textbf{ours}) & 2  & 37.0\% & 62.4\% & 77.5\% & \underline{10.0\%} & 3.83 & 27.7 & 0.174 & 15.5 \\
\ourproposedmethod (\textbf{ours}) & 5  & 37.2\% & 63.3\% & 78.4\% & ~~\textbf{9.8\%} & 3.68 & 27.1 & \underline{0.170} & 15.6 \\
\ourproposedmethod (\textbf{ours}) & 10 & 37.3\% & \textbf{63.7\%} & \underline{78.5\%} & ~~\textbf{9.8\%} & \underline{3.66} & \underline{27.0} & \textbf{0.169} & 16.3 \\
\ourproposedmethod (\textbf{ours}) & 20 & 37.4\% & \underline{63.4\%} & \underline{78.5\%} & ~~\textbf{9.8\%} & \textbf{3.64} & \textbf{26.9} & \textbf{0.169} & 17.0 \\
\ourproposedmethod (\textbf{ours}) & 50 & \textbf{37.5\%} & 63.3\% & \textbf{78.6\%} & ~~\textbf{9.8\%} & \textbf{3.64} & \textbf{26.9} & \textbf{0.169} & 20.3 \\
\bottomrule
\end{tabular}
\caption{Effect of the maximum number of references (\#Ref) on distance estimation performance and inference latency (T).}
\label{tab:latency}
\end{table*}

% \begin{table*}[t]
% \renewcommand\arraystretch{0.8}
% \centering
% \begin{tabular}{l|c|ccc|cccc}
% \toprule
% Method & \lidar{} & $\fivepercent$ & $\tenpercent$ & $\fifteenpercent$ & $\absrel$ & $\sqrel$ & $\rmse$ & $\rmselog$ \\
% \midrule
% SVR~\cite{gokcce2015vision} & \xmark & 34.1\% & 59.6\% & 76.6\% & 9.8\% & 0.99 & 7.5 & 0.133 \\
% DisNet~\cite{haseeb2018disnet} & \xmark & 34.2\% & 60.9\% & 77.7\% & 9.6\% & 0.93 & 7.3 & 0.129 \\
% Zhu~\etal~\cite{zhu2019learning} & \xmark & 61.8\% & 84.7\% & 93.5\% & 5.2\% & 0.31 & 4.2 & 0.074  \\
% Zhu~\etal~\cite{zhu2019learning} & \cmark & \textbf{62.9\%} & 84.6\% & 93.5\% & 5.2\% & 0.32 & 4.3 & 0.074  \\
% \Ourproposedmethod{} (\textbf{ours}) & \cmark & 62.7\% & \textbf{85.9\%} & \textbf{94.3\%} & \textbf{5.1\%} & \textbf{0.29} & \textbf{4.1} & \textbf{0.072} \\
% \bottomrule
% \end{tabular}
% \caption{Results on the pseudo long-range Waymo Open Dataset}
% \label{tab:main_results}
% \end{table*}

\begin{table*}[t]
\renewcommand\arraystretch{0.8}
\setlength\tabcolsep{4pt}
\small
\centering
\begin{tabular}{l|c|ccc|cccc}
\toprule
Method & \lidar{} & $\fivepercent$ & $\tenpercent$ & $\fifteenpercent$ & $\absrel$ & $\sqrel$ & $\rmse$ & $\rmselog$ \\
\midrule
SVR & \xmark & 33.0\% & 58.0\% & 74.0\% & 10.9\% & 1.56 & 10.7 & 0.166 \\
DisNet & \xmark & 29.5\% & 58.6\% & 75.0\% & 10.7\% & 1.46 & 10.5 & 0.162 \\
\cite{zhu2019learning} & \xmark & 40.3\% & 66.7\% & 80.3\% & ~~8.4\% & 0.91 & ~~8.6 & 0.121  \\
\cite{zhu2019learning} & \cmark & 37.7\% & 63.5\% & 77.2\% & ~~9.2\% & 1.06 & ~~9.2 & 0.132  \\
\Ourproposedmethod{} (\textbf{ours}) & \cmark & \textbf{44.2\%} & \textbf{71.1\%} & \textbf{84.6\%} & \textbf{~~7.6\%} & \textbf{0.75} & \textbf{~~7.7} & \textbf{0.110} \\
\bottomrule
\end{tabular}
\caption{Results on the pseudo long-range nuScenes Dataset. \Ourproposedmethod{} outperforms SVR~\citep{gokcce2015vision}, DisNet~\citep{haseeb2018disnet}, and \citet{zhu2019learning}.}
\label{tab:nuscenes_results}
\end{table*}

% \section{Results on Other Derived Pseudo Long-Range Datasets}

% We further provide two derived pseudo long-rang datasets, \textit{Pseudo Long-Range Waymo Open Dataset} and \textit{Pseudo Long-Range nuScenes Dataset}, and the experimental results on them. Although we prove several derived pseudo long-range dataset, we still encourage readers to focus more on our \longrangdataset{} dataset. The proposed \longrangdataset{} dataset is a large-scale dataset with annotation for \textit{real} long-range objects, which aligns better with the real-world scenario.

% \paragraph{Pseudo Long-Range Waymo Open Dataset.} Similar to constructing the pseudo long-range KITTI dataset, we modify the Waymo Open Dataset resulting in the pseudo long-range Waymo Open Dataset. We assume \lidar{} has an effective sensing range of 40 meters only. We remove \lidar{} points beyond 40 meters, and objects lying beyond this range are considered as long-range target objects. The dataset includes 121,769 training images containing 403,305 vehicles and 29,939 validation images with 99,463 vehicles. Results are shown in Table~\ref{tab:main_results}. Zhu~\etal~\cite{zhu2019learning} achieves good performance, likely due to the fact that the data distribution of training and validation set is similar in Waymo Open Dataset and the target vehicles are not actual long-range objects. \Ourproposedmethod{} further improves this strong baseline. For example, with \ourproposedmethod{}, the $\tenpercent$ metric increases by 1.2\% (from 84.7\% to 85.9\%).

\section{Pseudo Long-Range nuScenes Dataset} 
We further provide another derived pseudo long-rang dataset, \textit{Pseudo Long-Range nuScenes Dataset}, and the experimental results on it. Although we prove several derived pseudo long-range datasets, we still encourage readers to focus more on our \longrangdataset{} dataset. The proposed \longrangdataset{} dataset is a large-scale dataset with annotation for \textit{real} long-range objects, which aligns better with the real-world scenario. 

To construct pseudo long-range nuScenes dataset, we first export the original nuScenes dataset to KITTI format by the nuScenes offical code\footnote{\url{https://github.com/nutonomy/nuscenes-devkit/blob/master/python-sdk/nuscenes/scripts/export_kitti.py}}. Then we reuse the pipeline that builds the pseudo long-range KITTI dataset. The final pseudo long-range nuScenes dataset includes 18,926 training images with target 59,800 vehicles, and 4,017 validation images with 11,737 target vehicles.

The results are shown in Table~\ref{tab:nuscenes_results}. \Ourproposedmethod{} outperforms all baselines with a large margin. For example, with \ourproposedmethod{}, \citet{zhu2019learning} (line 3) improves by 4.4\% on the $\tenpercent$ metric (from 66.7\% to 71.1\%). By comparing line 3 with line 4, we observe the performance drop when appending the \lidar{} points as an additional input channel to \citet{zhu2019learning}. This observation also supports our conclusion in the main manuscript, the additional distance channel does not help to improve the distance estimation performance. The observation here is even more obvious, because the original nuScenes dataset only provides relative low resolution (32 beams, while KITTI dataset is
64 beams) \lidar{} signals~\citep{nuscenes2019}.

\begin{table*}
\renewcommand\arraystretch{0.8}
\centering
\small
\begin{tabular}{l|c|ccc|cccc}
\toprule
Method & \lidar{} & $\fivepercent$ & $\tenpercent$ & $\fifteenpercent$ & $\absrel$ & $\sqrel$ & $\rmse$ & $\rmselog$ \\
\midrule
SMOKE & \xmark & 42.4\% & 70.5\% & 83.3\% & 10.3\% & 1.98 & 10.6 & 0.145  \\
RTM3D & \xmark & 42.7\% & 71.1\% & \textbf{83.9\%} & ~~9.8\% & 1.75 & ~~9.9 & 0.135  \\
\Ourproposedmethod{} (\textbf{ours}) & \cmark & \textbf{46.3\%} & \textbf{72.5\%} & \textbf{83.9\%} & \textbf{~~7.5\%} & \textbf{0.68} & \textbf{~~6.8} & \textbf{0.112}  \\
\bottomrule
\end{tabular}
\caption{Compare with monocular 3D object detection methods on pseudo long-range KITTI dataset. \Ourproposedmethod{} outperforms SMOKE~\citep{liu2020smoke}, and RTM3D~\citep{li2020rtm3d}.}
\label{tab:compare_detection}
\end{table*}

\section{Compare with Monocular 3D Object Detectors}
In the related work section of the main manuscript, we have discussed the relationship between our proposed task, namely Long-Range Distance Estimation, and monocular 3D object detection. Here, we conduct experiments to compare \ourproposedmethod{} to monocular 3D detection methods on pseudo long-range KITTI dataset. We associate the detected 3D bounding boxes with 2D ground-truth boxes using a bipartite matching algorithm. The weights of the bipartite graph are the 2D IoUs of ground-truth and detected boxes. In order to compare with our method, we calculate the distance prediction metrics on matched ground-truth and detected box pairs. 

As shown in Table~\ref{tab:compare_detection}, \ourproposedmethod{} outperforms SOTA 3D object detectors on the distance prediction task. For example, on the $\fivepercent$ metric, \ourproposedmethod{} improves 3.6\% (from 42.7\% to 46.3\%) compared with previous state-of-the-art monocular 3D object detector RMT3D~\citep{li2020rtm3d}. This improvement is more significant than the improvement from SMOKE~\citep{liu2020smoke} to RTM3D~\citep{li2020rtm3d} (only improves 0.3\% on the $\fivepercent$ metric) due to \ourproposedmethod{} utilizing the short-range \lidar{} signals. It is also worth noting that \ourproposedmethod{} does not utilized the techniques proposed in other monocular 3D object detection papers such as multi-task training and strong architectures. Therefore, \ourproposedmethod{} is complementary to 3D detection methods, and combining it with such methods will further improve the long-range distance estimation performance. 

\section{More Qualitative Examples on \LongRangDataset{} Dataset} \label{sec:vis_attention}
We also provide more qualitative examples of the output of the attention module in the supplemental material (please download the zip file). All images are from our proposed \longrangdataset{} dataset. Similar to Figure~\ref{fig:vis_attention} in the main manuscript, the target objects are highlighted in yellow. References with the largest attention weights are highlighted in red. Other references are colored in blue. As before, we observe that the most important references are often close to or in the same lane as the target, and share a similar pose. 
\end{document}

\blfootnote{$^\ast$Work done while at Waymo LLC. \dag Equal contribution.}

\maketitle

%%%%%%%%% BODY TEXT
\appendix

\section*{More Qualitative Examples on \LongRangDataset{} Dataset} 
We also provide more qualitative examples of the output of the attention module in this file. All images are from our proposed \longrangdataset{} dataset. Similar to Figure 8 in the main manuscript, the target objects are highlighted in yellow. References with the largest attention weights are highlighted in red. Other references are colored in blue. As before, we observe that the most important references are often close to or in the same lane as the target, and share a similar pose. 

% \newpage

\begin{figure}[h]
\centering
\includegraphics[width=1\linewidth]{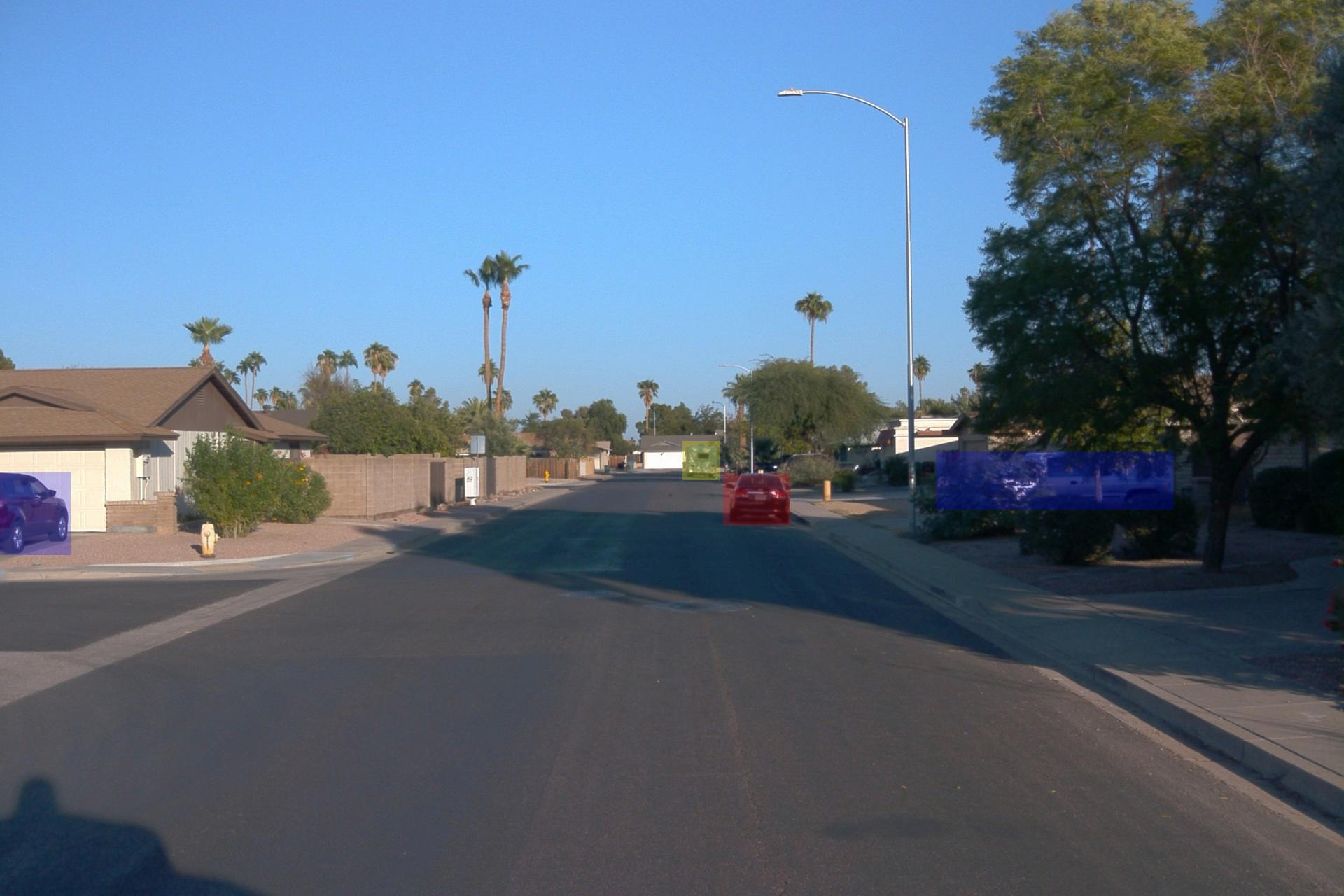}
\label{fig:vis_113_0}
\end{figure}

\begin{figure}[h]
\centering
\includegraphics[width=1\linewidth]{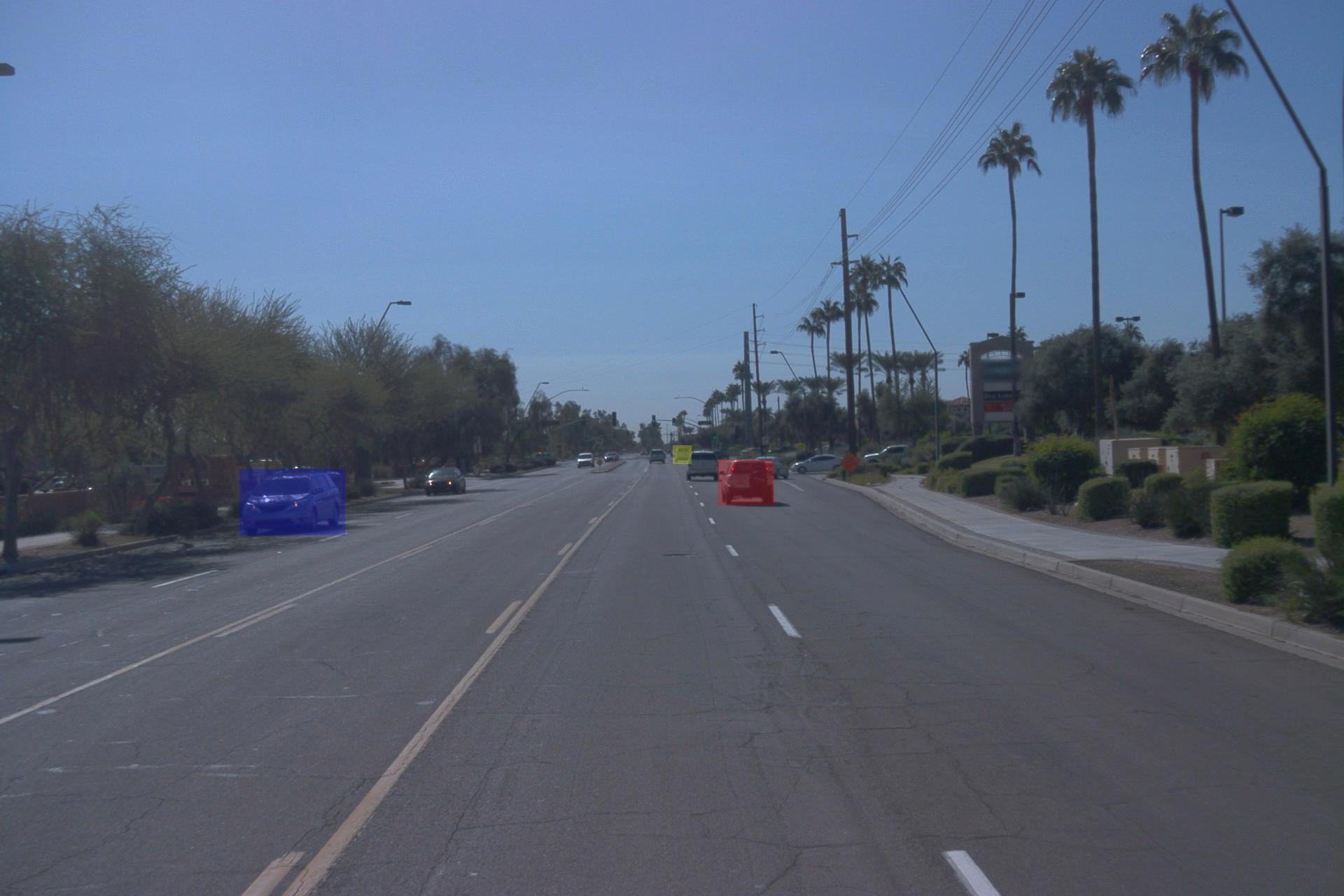}
\label{fig:vis_140_3}
\end{figure}

\begin{figure}[h]
\centering
\includegraphics[width=1\linewidth]{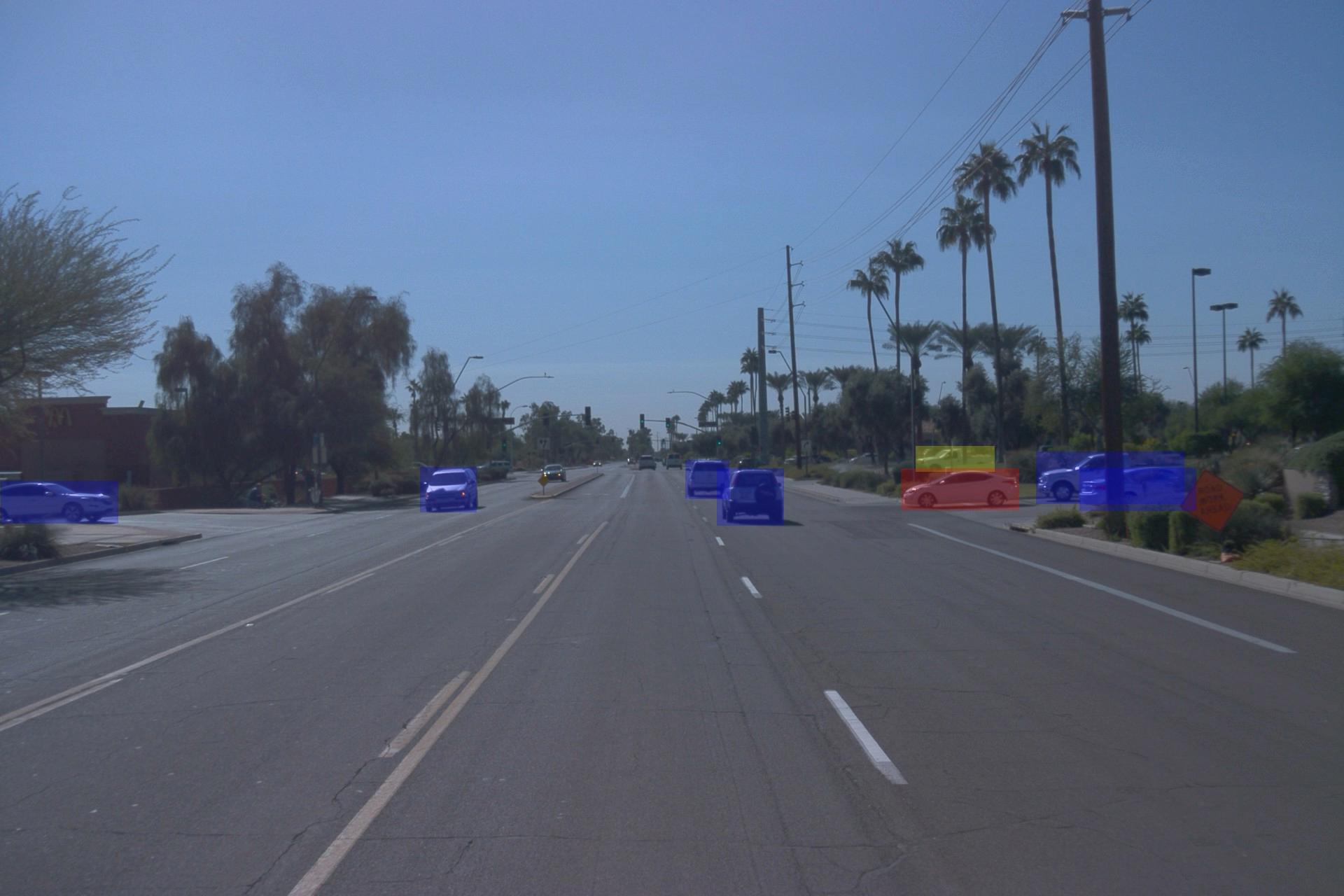}
\label{fig:vis_143_0}
\end{figure}

\begin{figure}[h]
\centering
\includegraphics[width=1\linewidth]{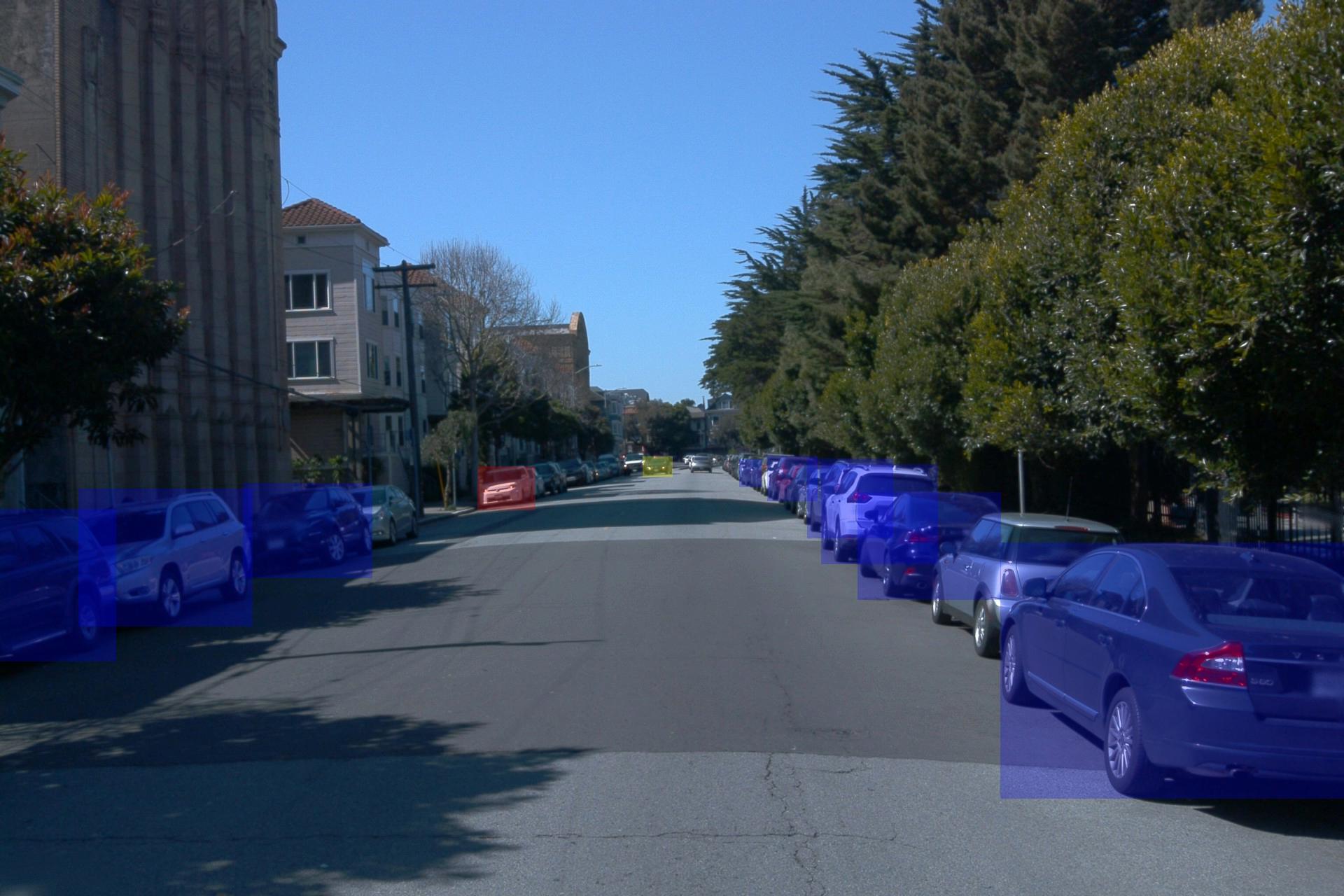}
\label{fig:vis_150_3}
\end{figure}

\begin{figure}[h]
\centering
\includegraphics[width=1\linewidth]{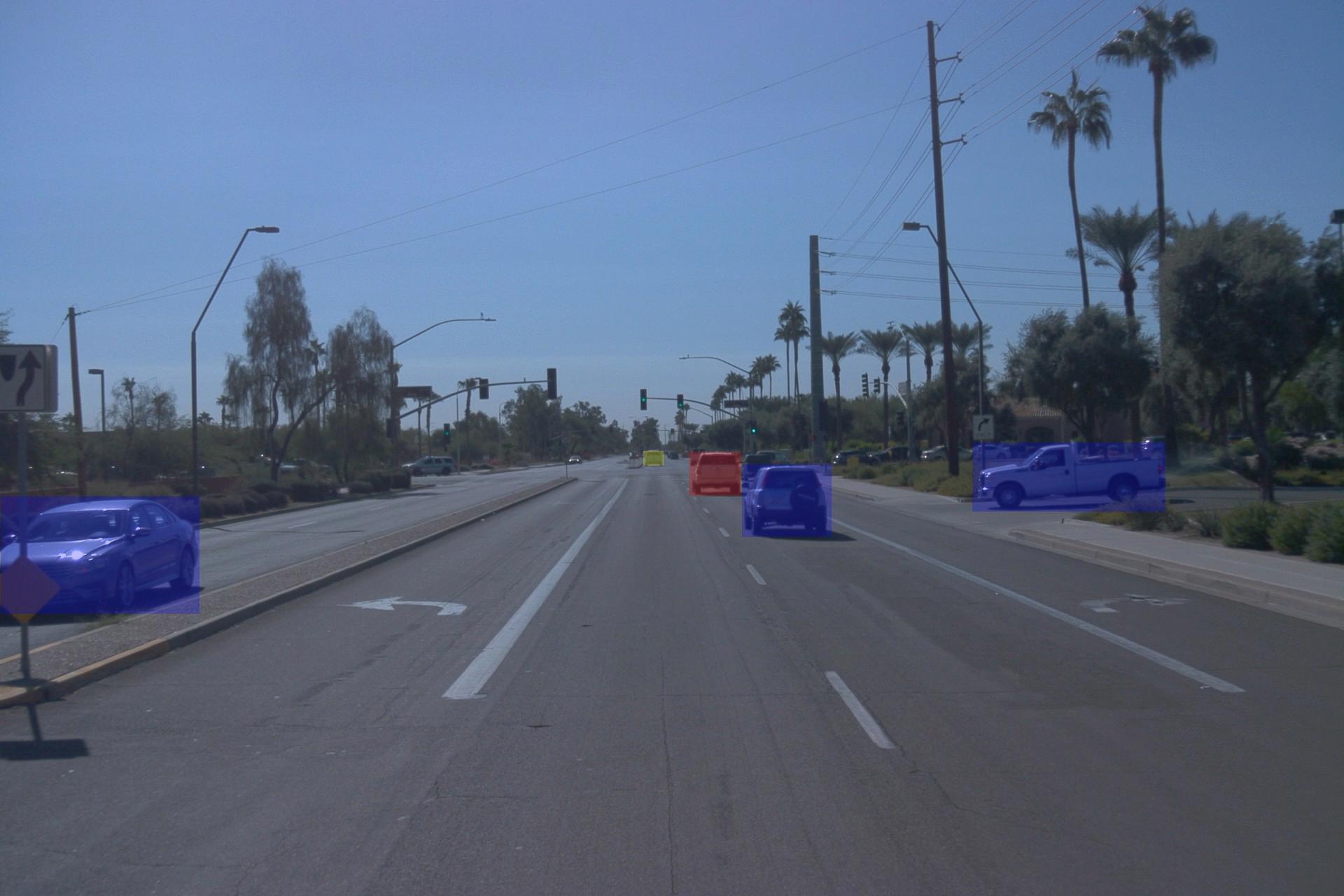}
\label{fig:vis_154_7}
\end{figure}

\begin{figure}[h]
\centering
\includegraphics[width=1\linewidth]{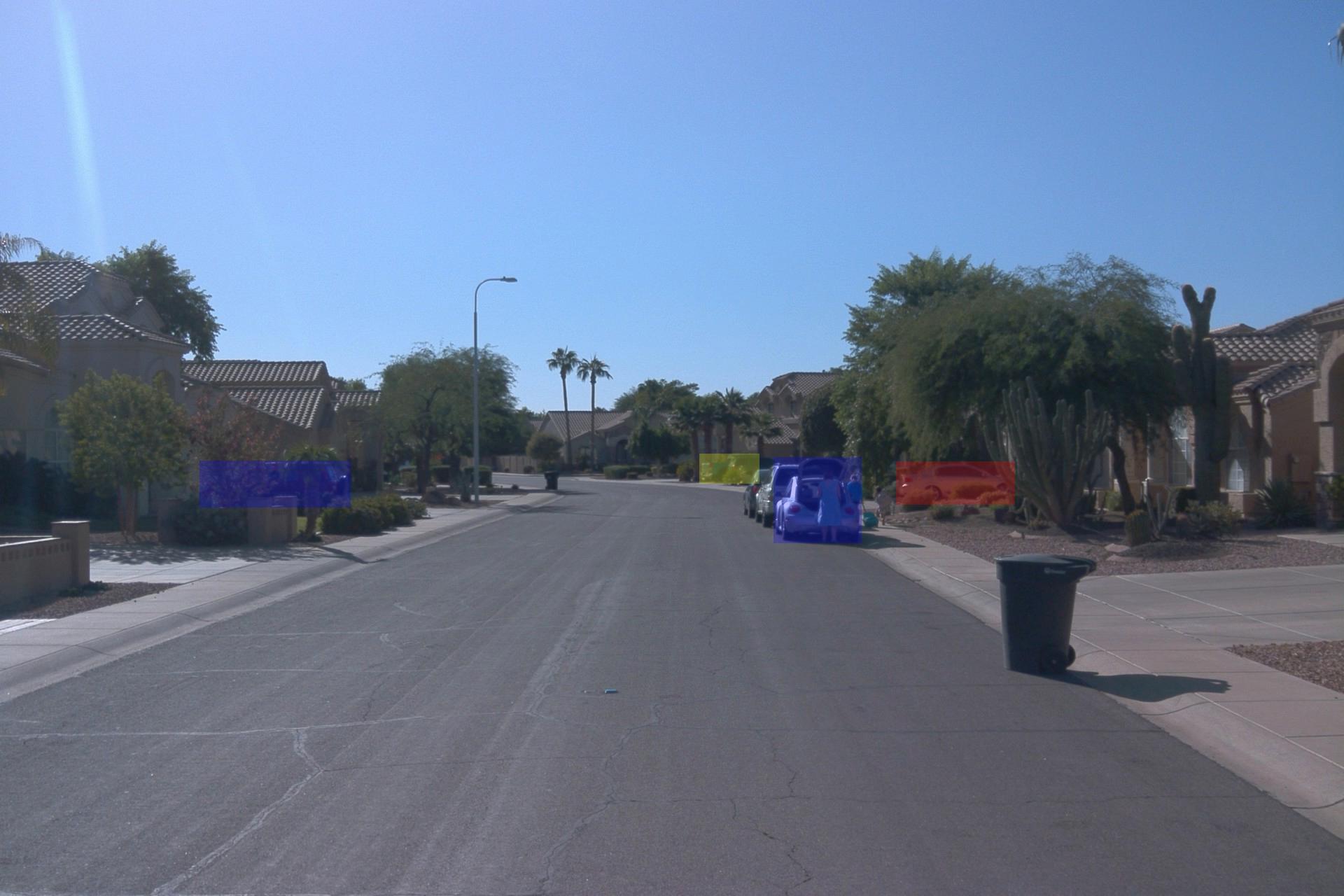}
\label{fig:vis_15_2}
\end{figure}

\begin{figure}[h]
\centering
\includegraphics[width=1\linewidth]{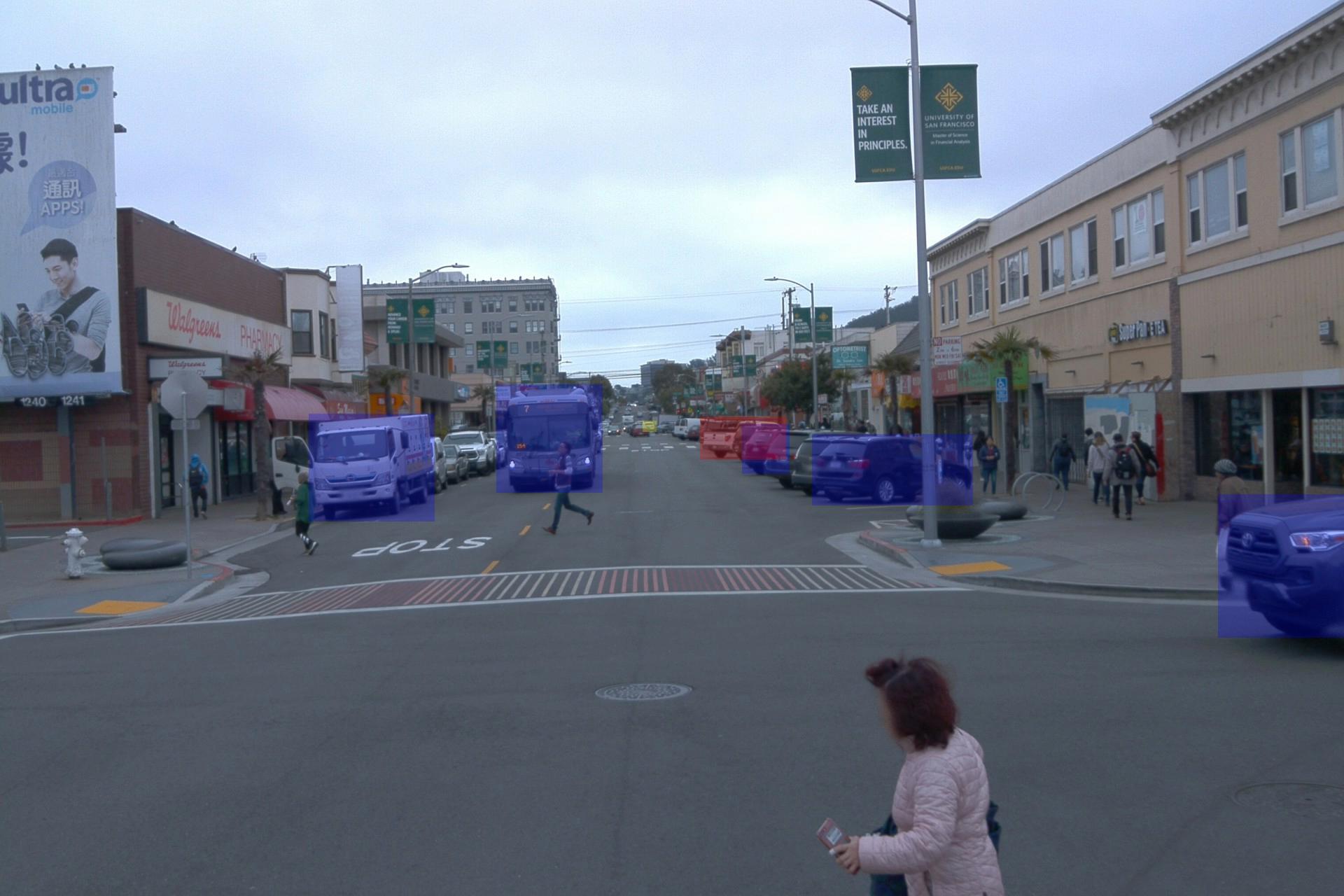}
\label{fig:vis_230_6}
\end{figure}

\begin{figure}[h]
\centering
\includegraphics[width=1\linewidth]{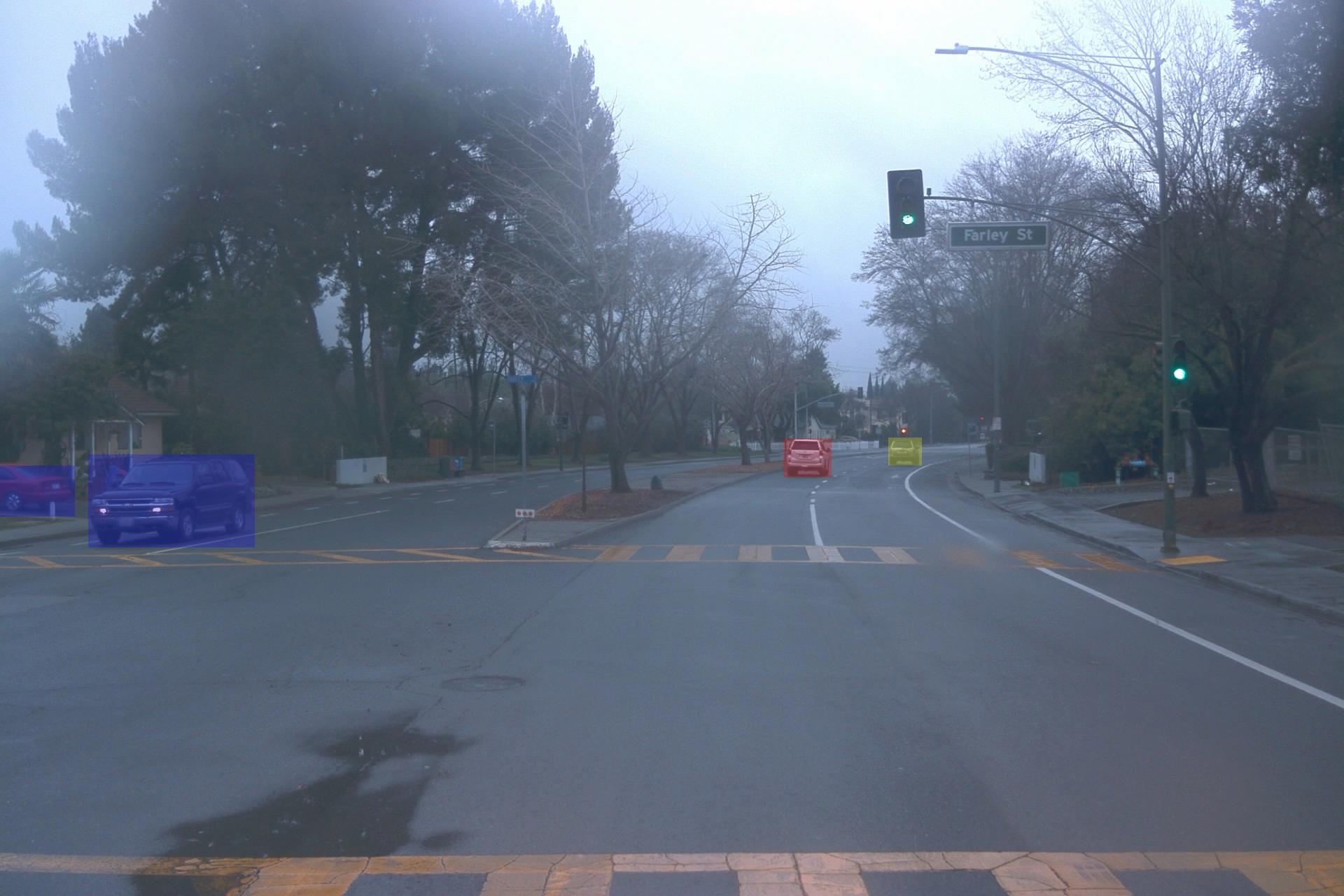}
\label{fig:vis_233_1}
\end{figure}

\begin{figure}[h]
\centering
\includegraphics[width=1\linewidth]{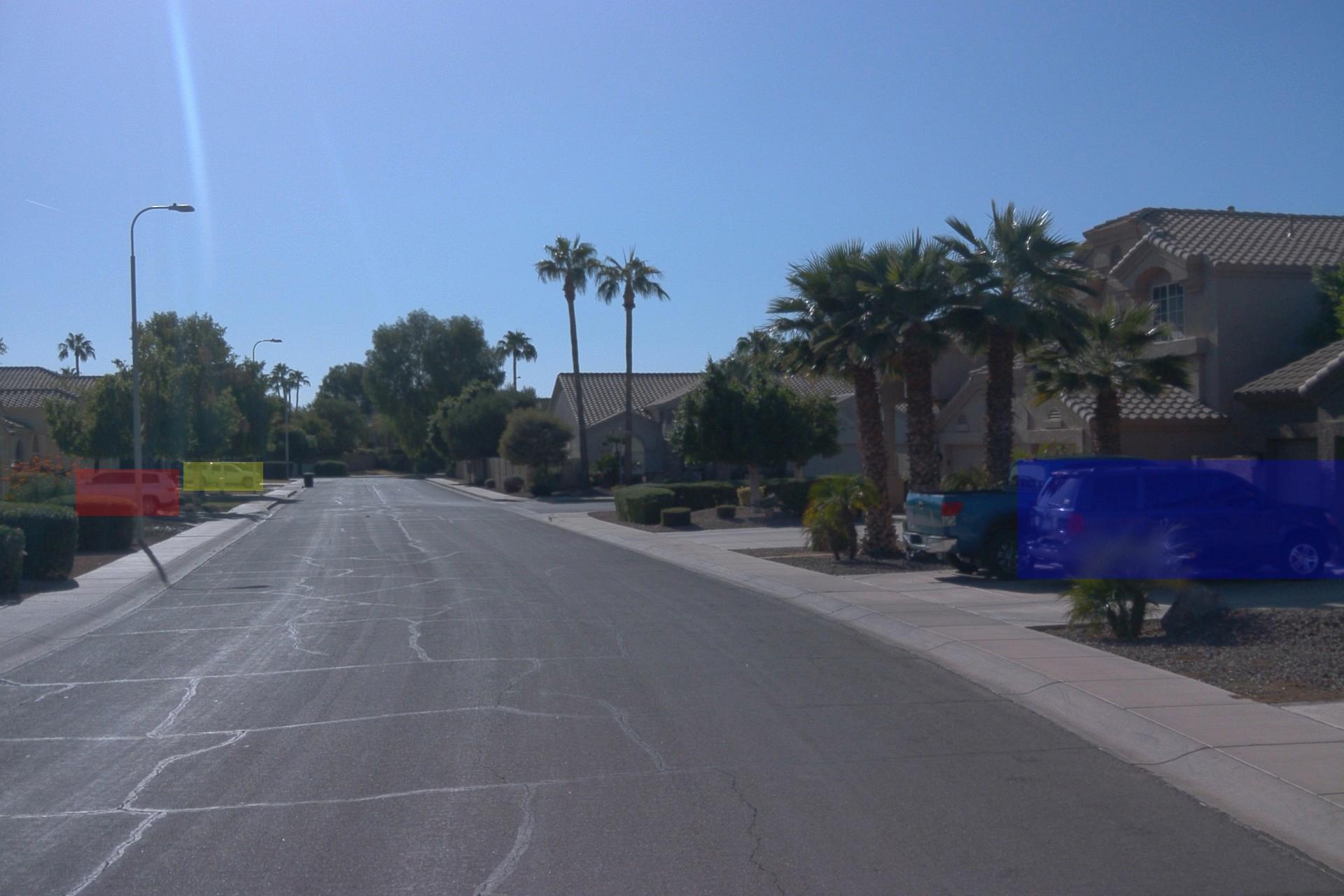}
\label{fig:vis_23_2}
\end{figure}

\begin{figure}[h]
\centering
\includegraphics[width=1\linewidth]{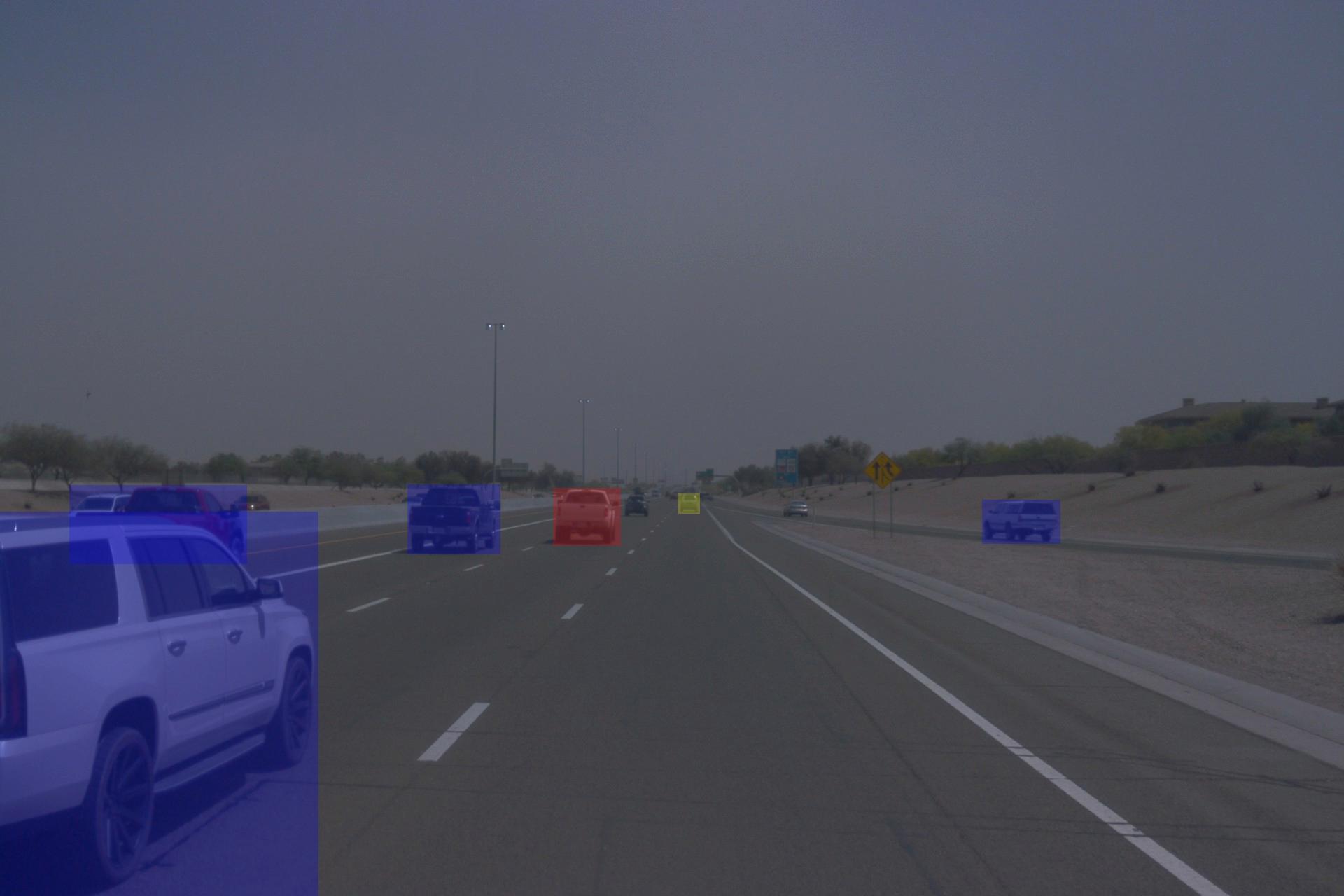}
\label{fig:vis_241_3}
\end{figure}

\begin{figure}[h]
\centering
\includegraphics[width=1\linewidth]{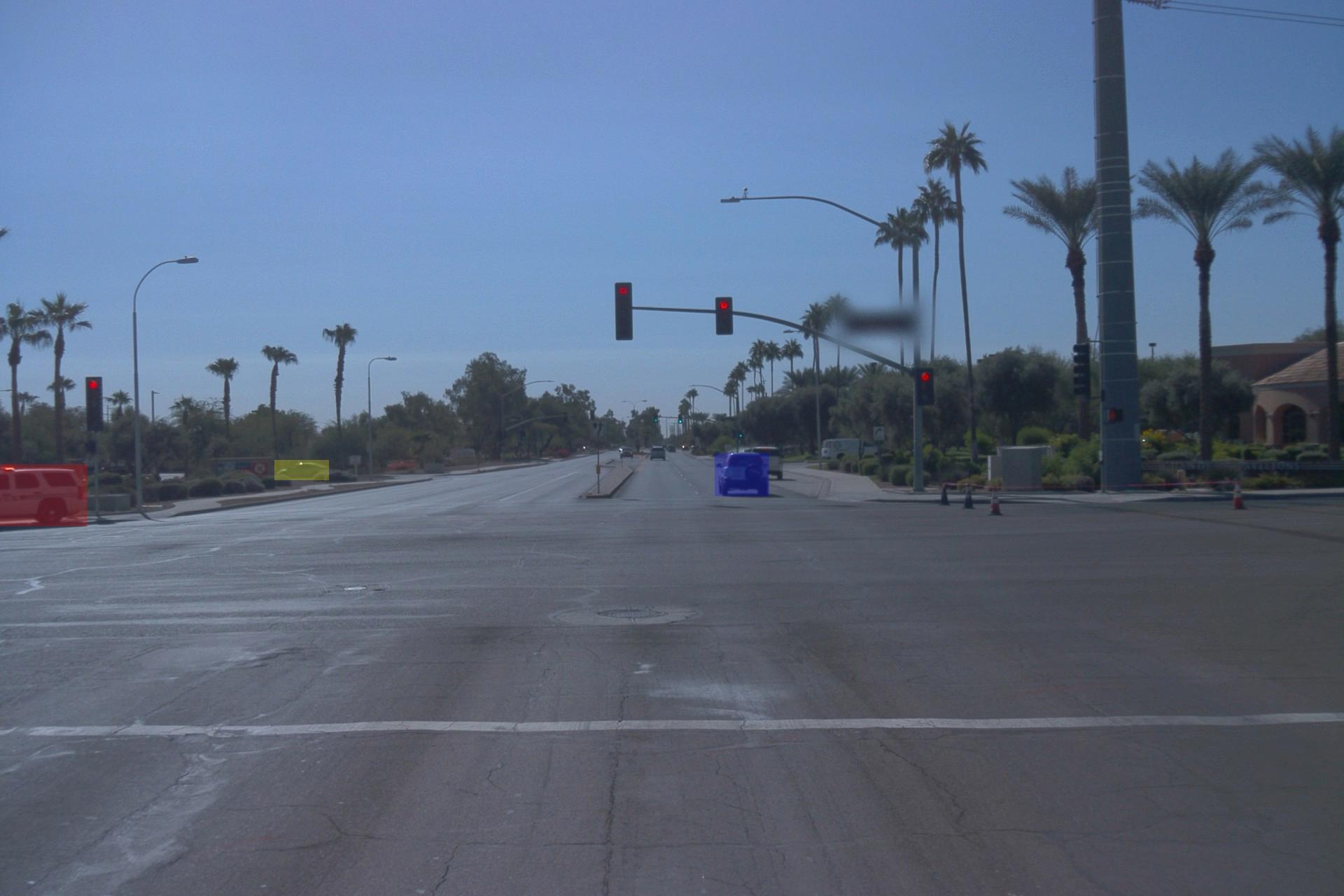}
\label{fig:vis_241_4}
\end{figure}

\begin{figure}[h]
\centering
\includegraphics[width=1\linewidth]{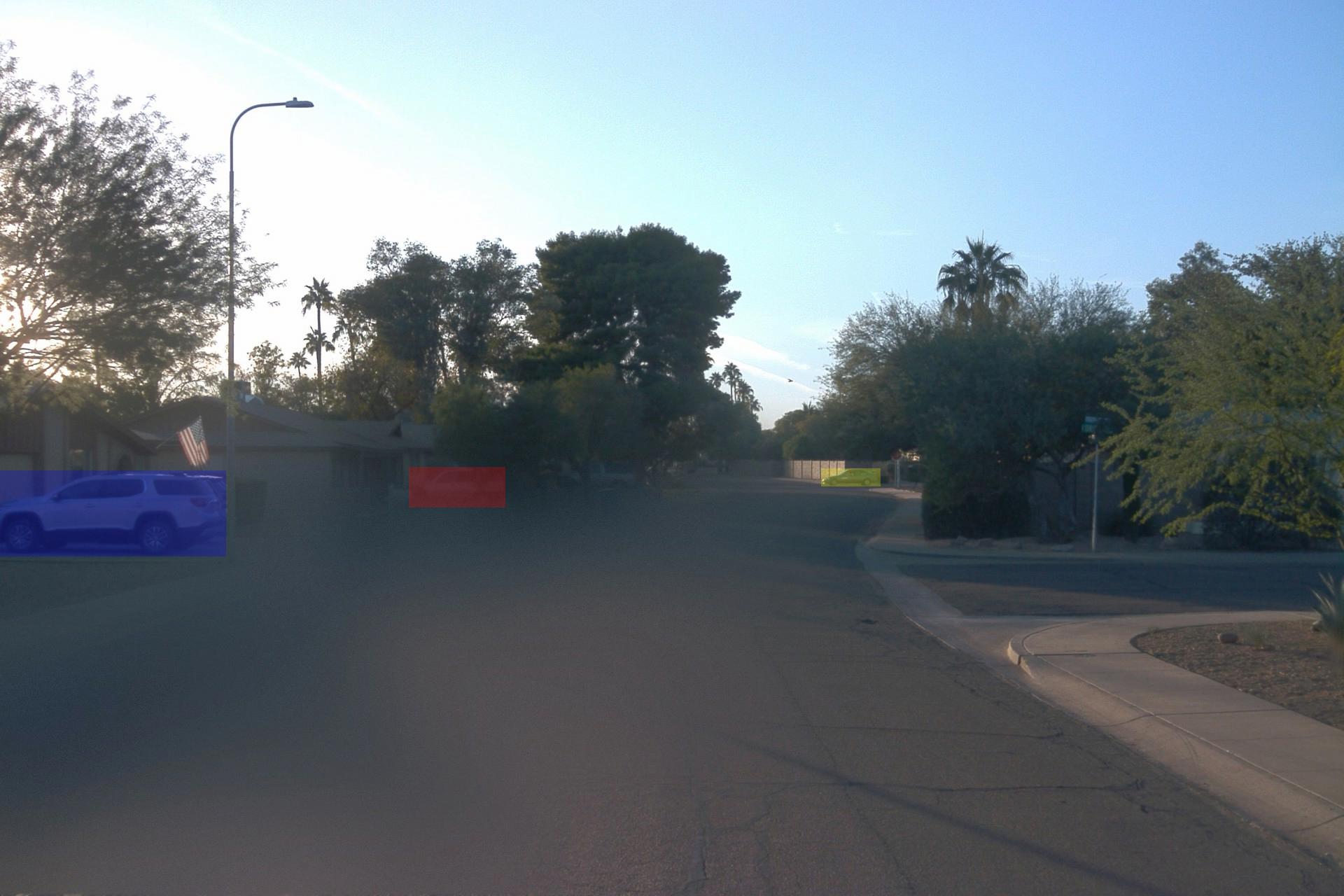}
\label{fig:vis_257_0}
\end{figure}

\begin{figure}[h]
\centering
\includegraphics[width=1\linewidth]{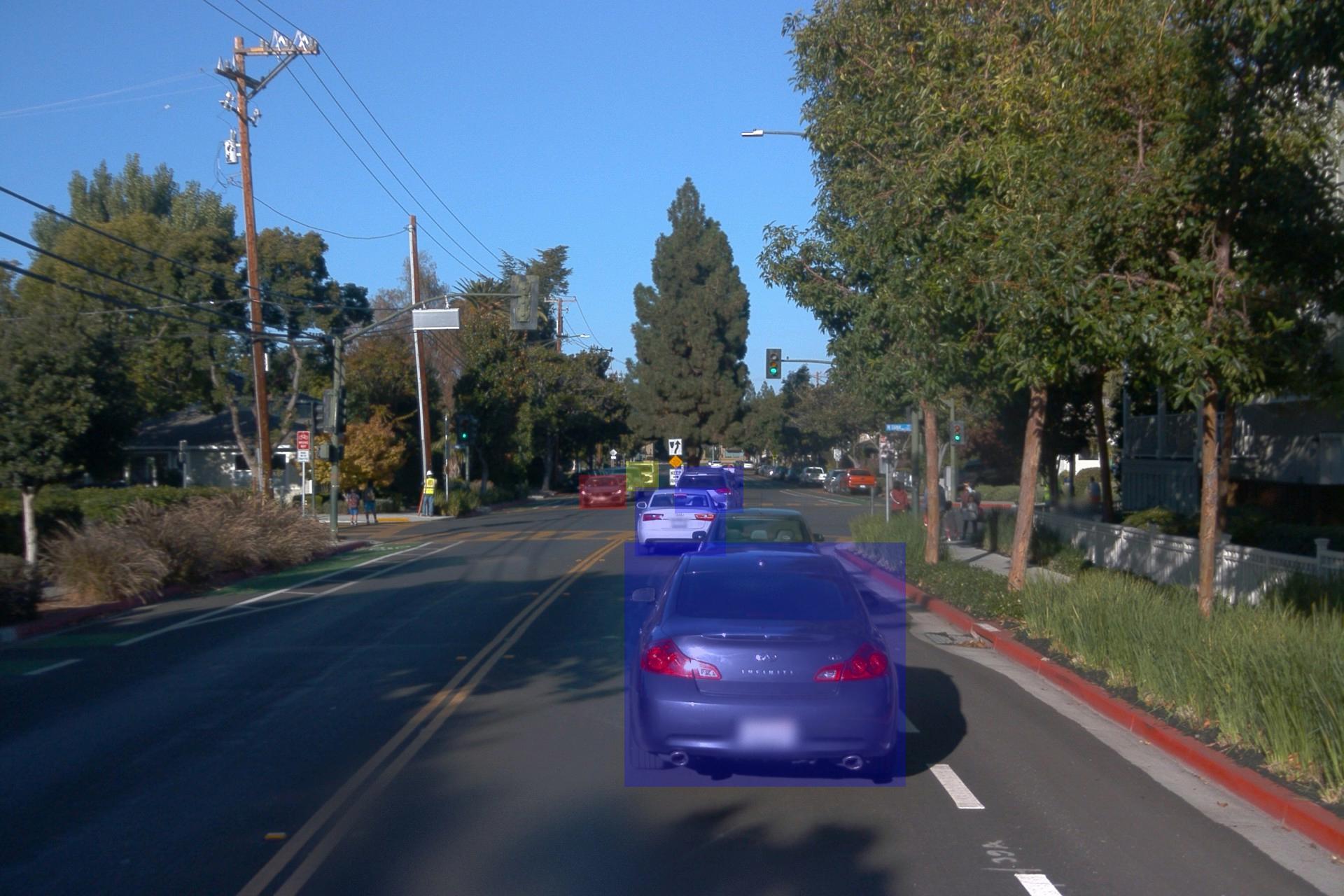}
\label{fig:vis_267_5}
\end{figure}

\begin{figure}[h]
\centering
\includegraphics[width=1\linewidth]{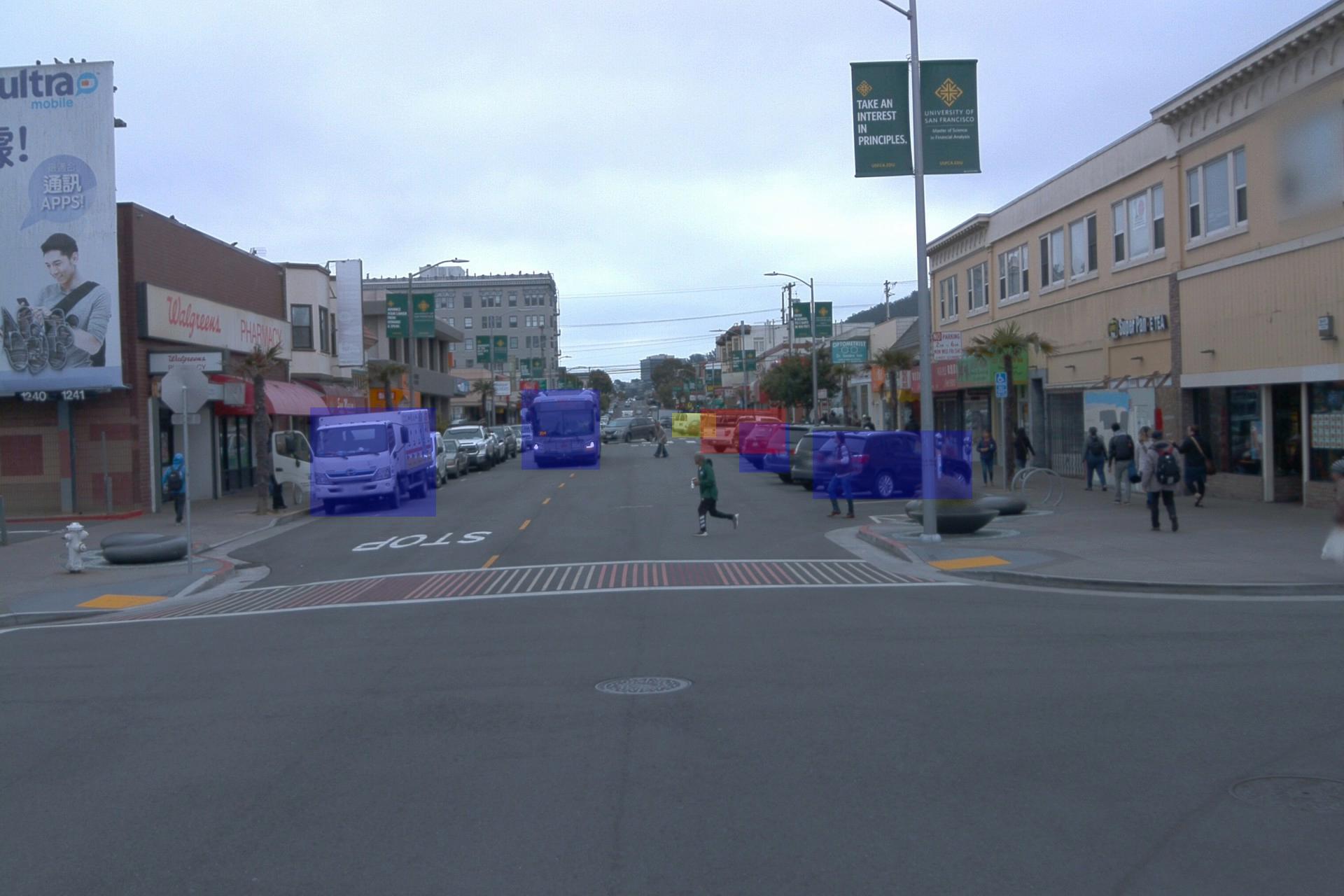}
\label{fig:vis_268_0}
\end{figure}

\begin{figure}[h]
\centering
\includegraphics[width=1\linewidth]{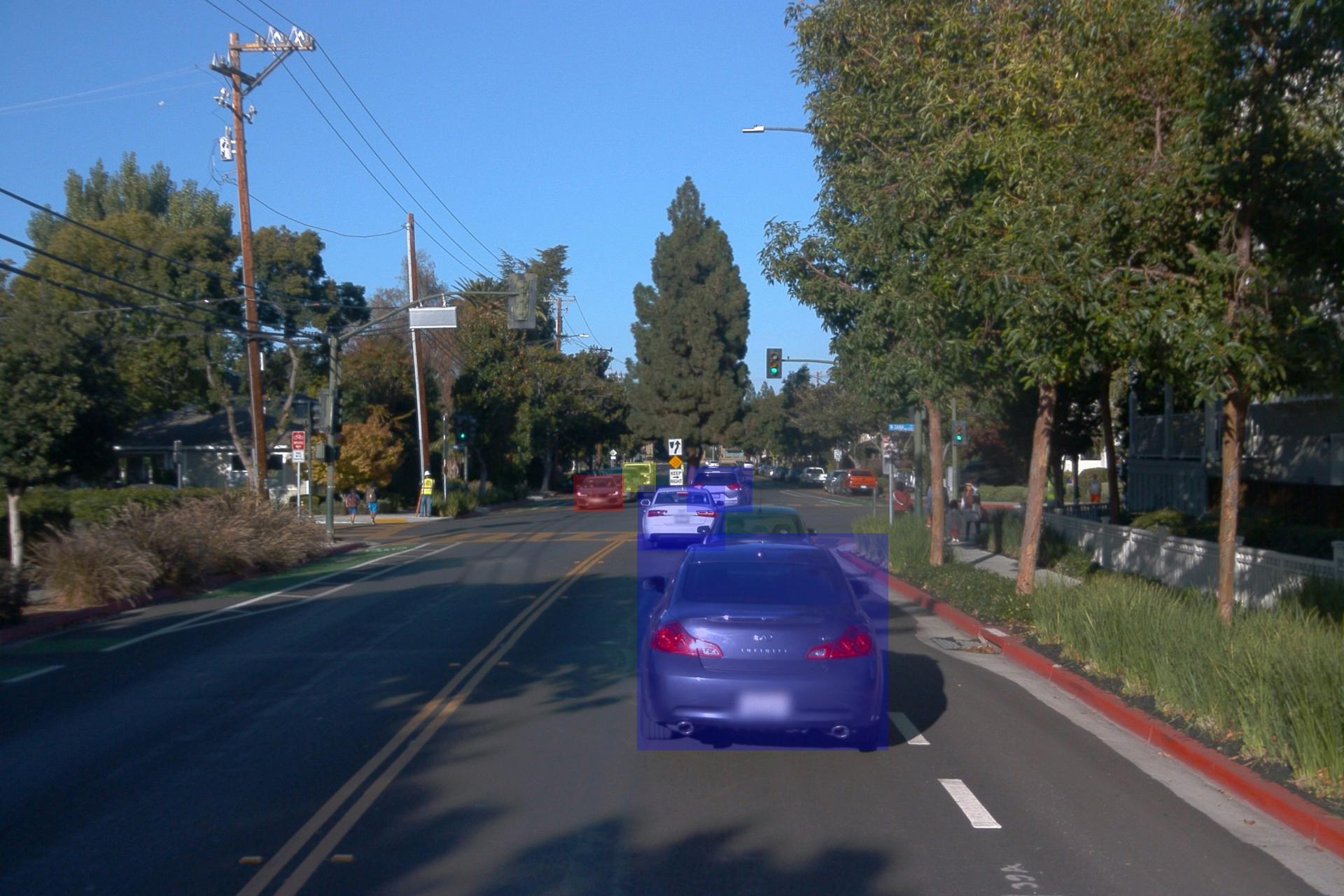}
\label{fig:vis_270_4}
\end{figure}

\begin{figure}[h]
\centering
\includegraphics[width=1\linewidth]{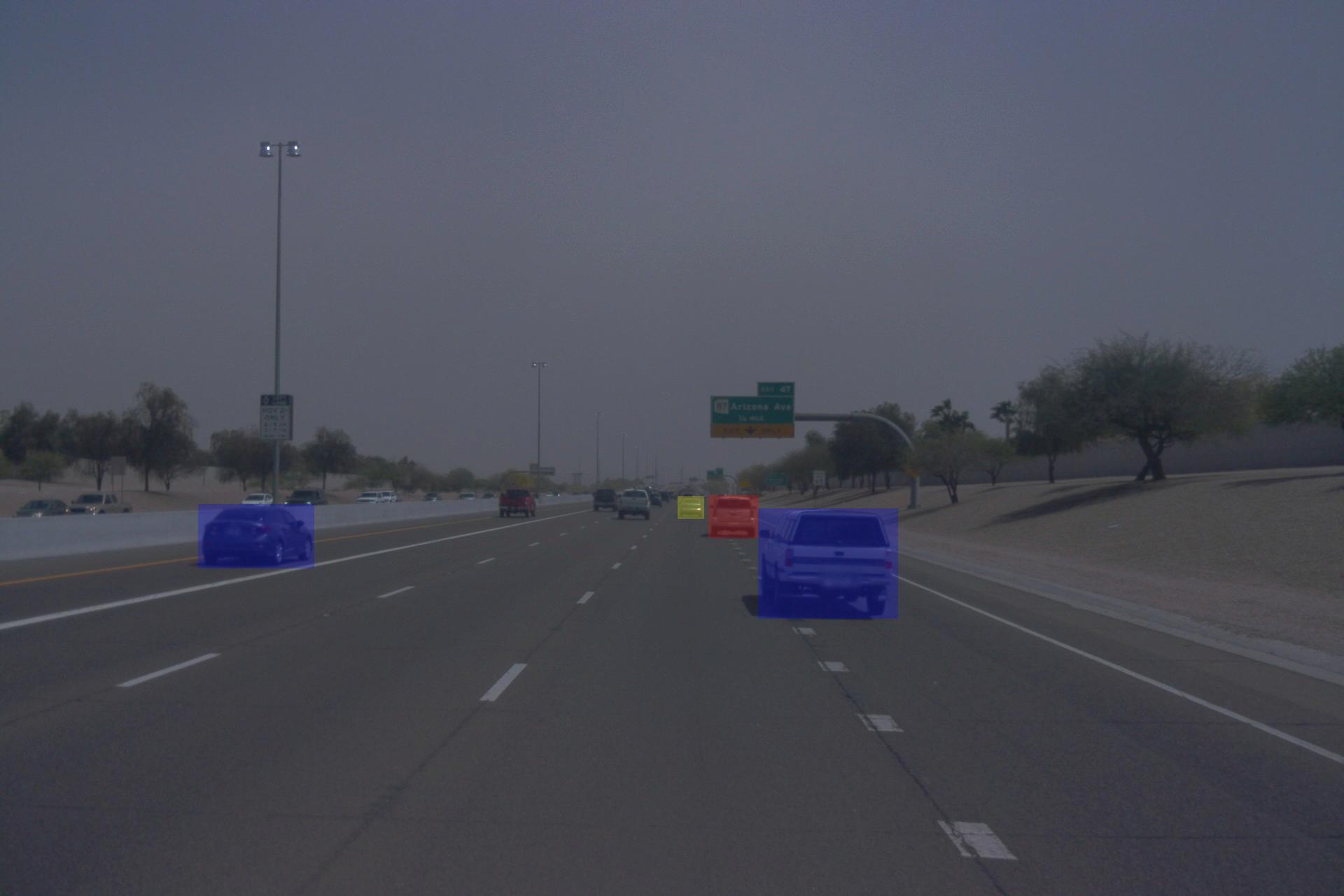}
\label{fig:vis_271_2}
\end{figure}

\begin{figure}[h]
\centering
\includegraphics[width=1\linewidth]{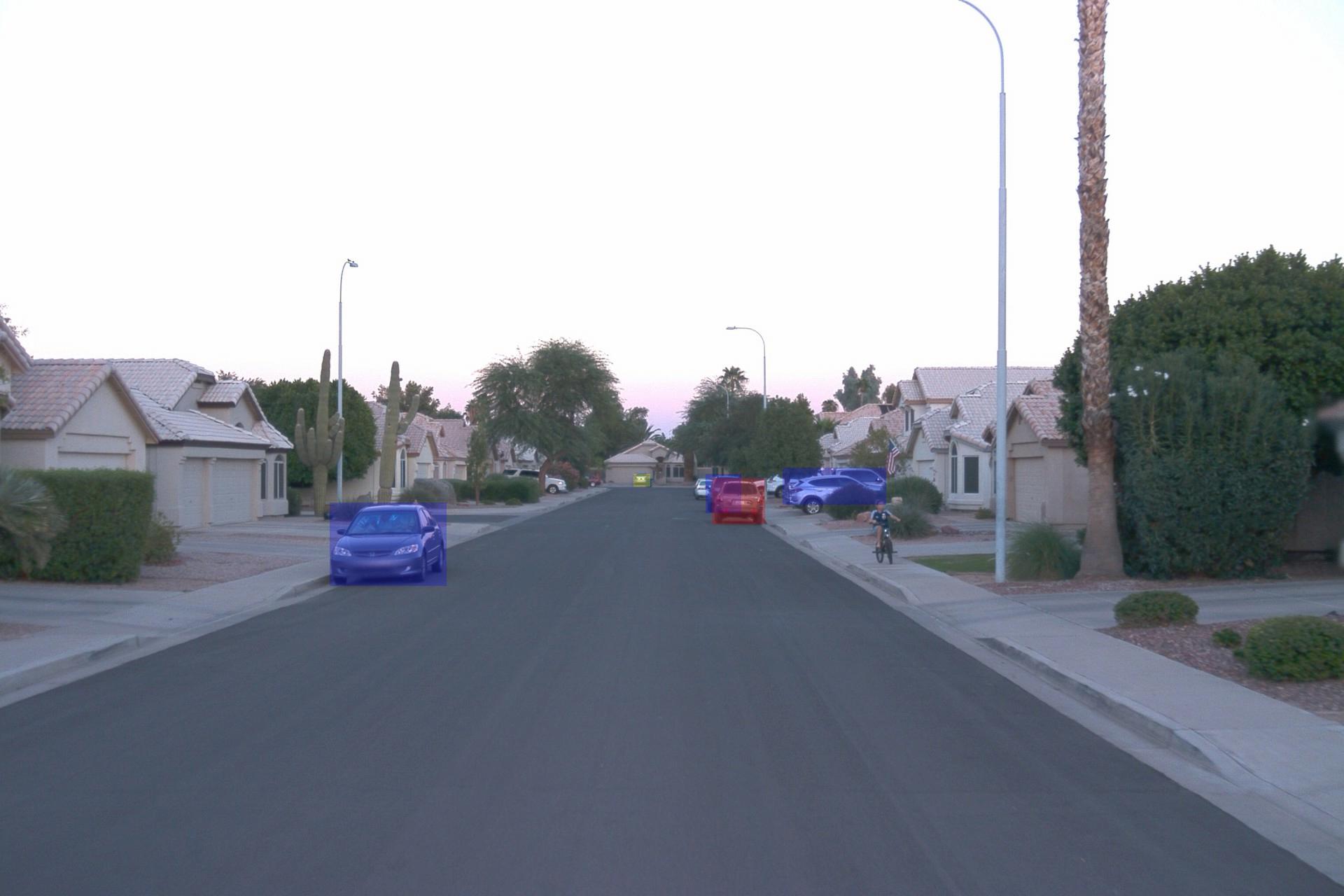}
\label{fig:vis_28_4}
\end{figure}

\begin{figure}[h]
\centering
\includegraphics[width=1\linewidth]{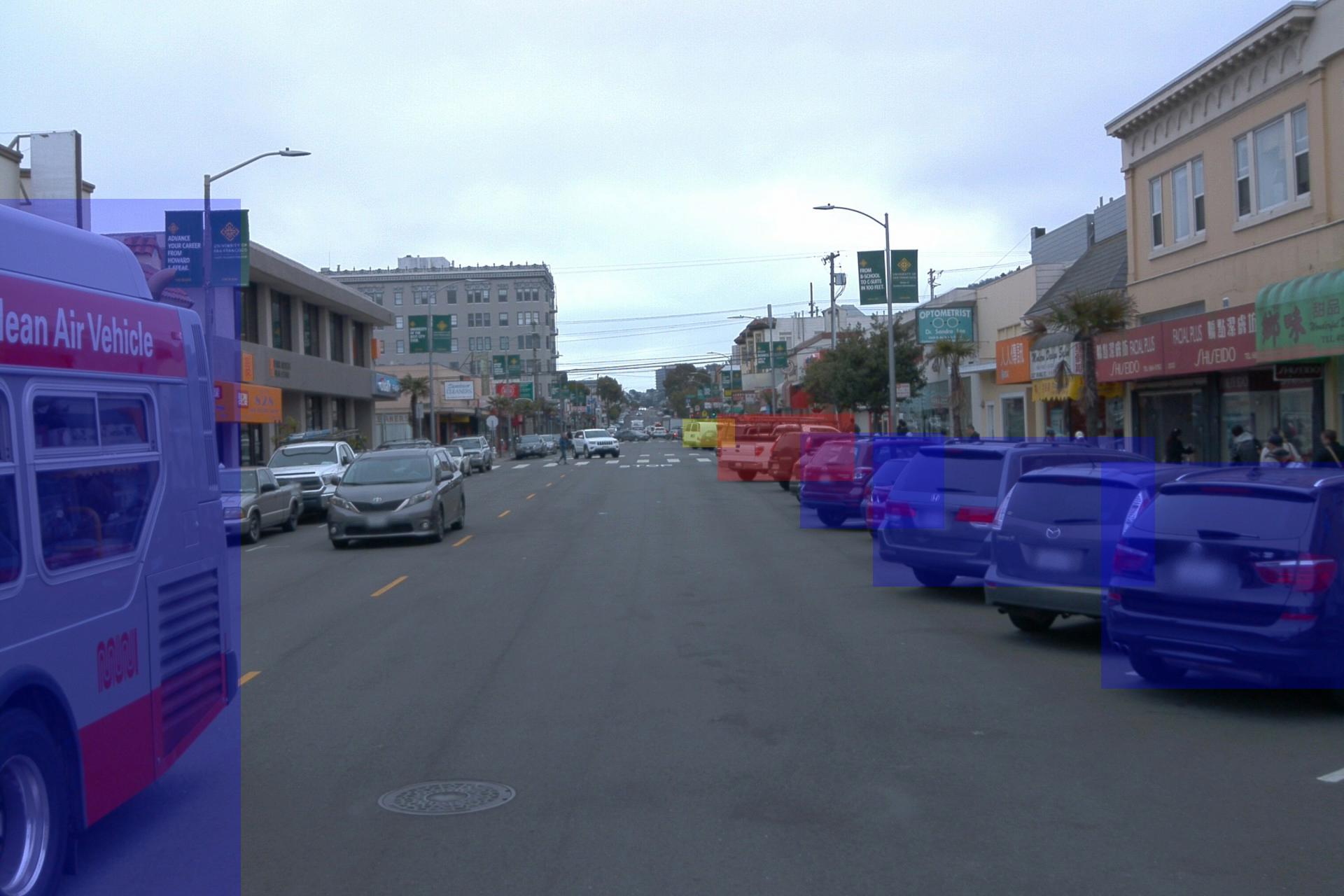}
\label{fig:vis_318_3}
\end{figure}

\begin{figure}[h]
\centering
\includegraphics[width=1\linewidth]{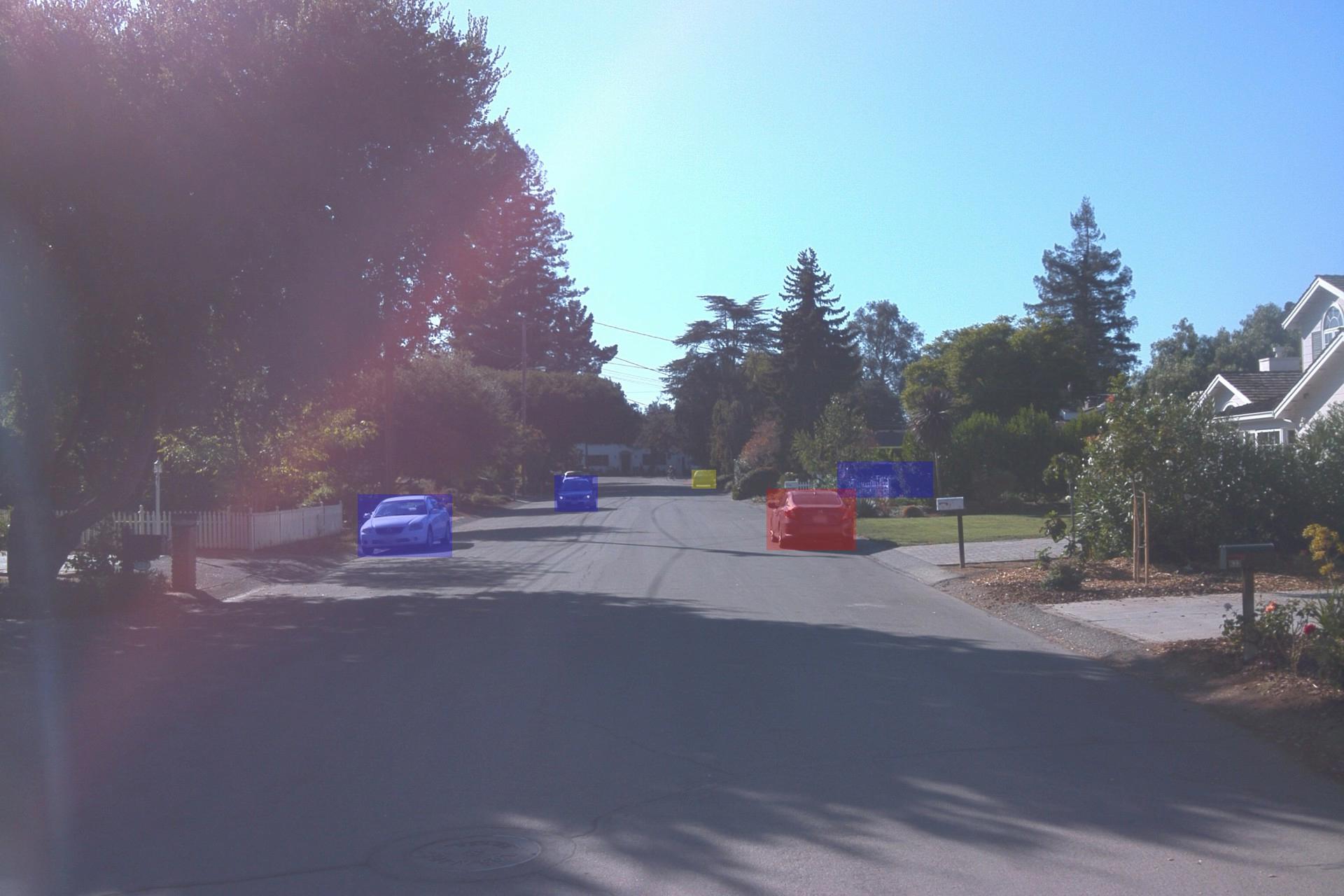}
\label{fig:vis_342_6}
\end{figure}

\begin{figure}[h]
\centering
\includegraphics[width=1\linewidth]{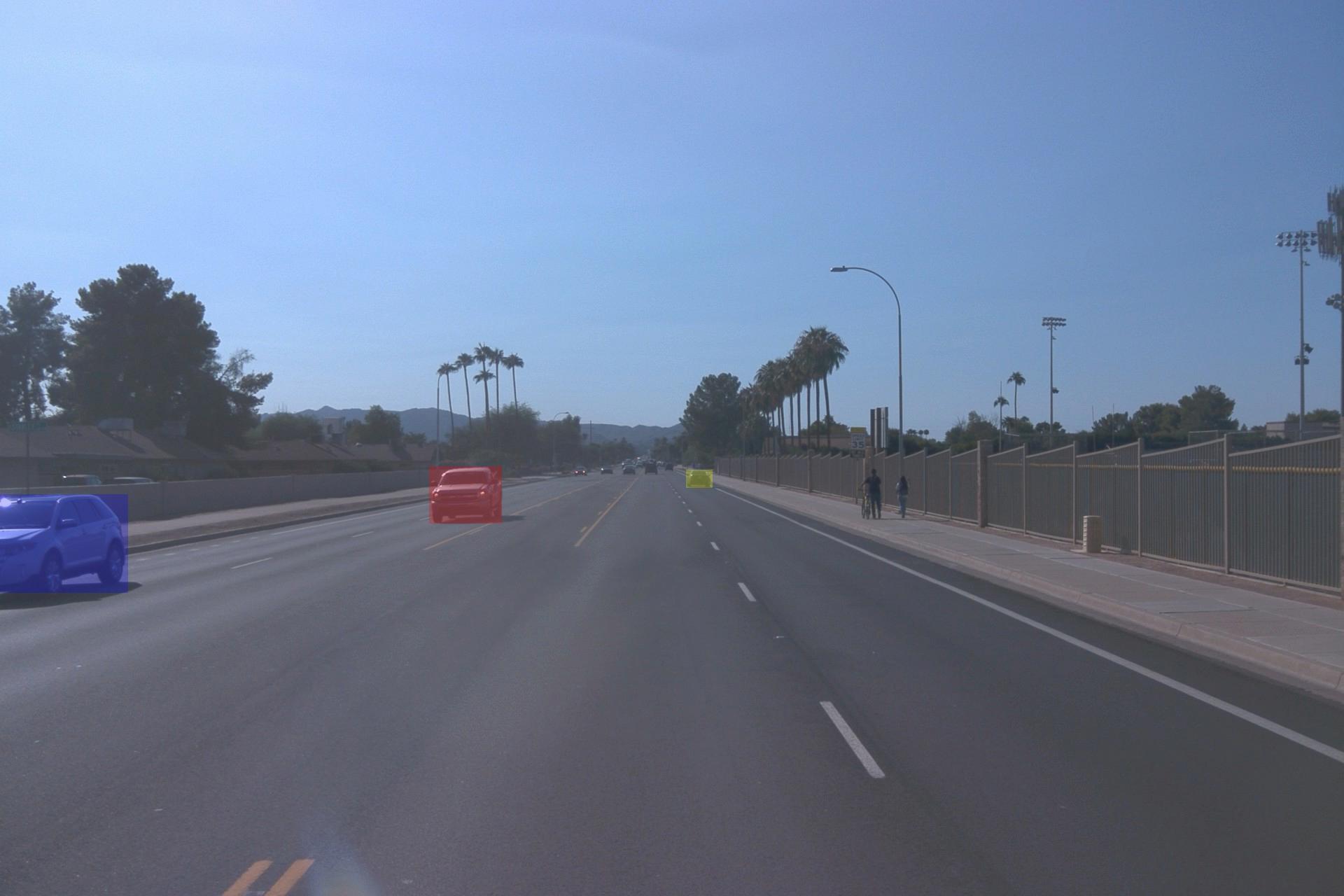}
\label{fig:vis_404_1}
\end{figure}

\begin{figure}[h]
\centering
\includegraphics[width=1\linewidth]{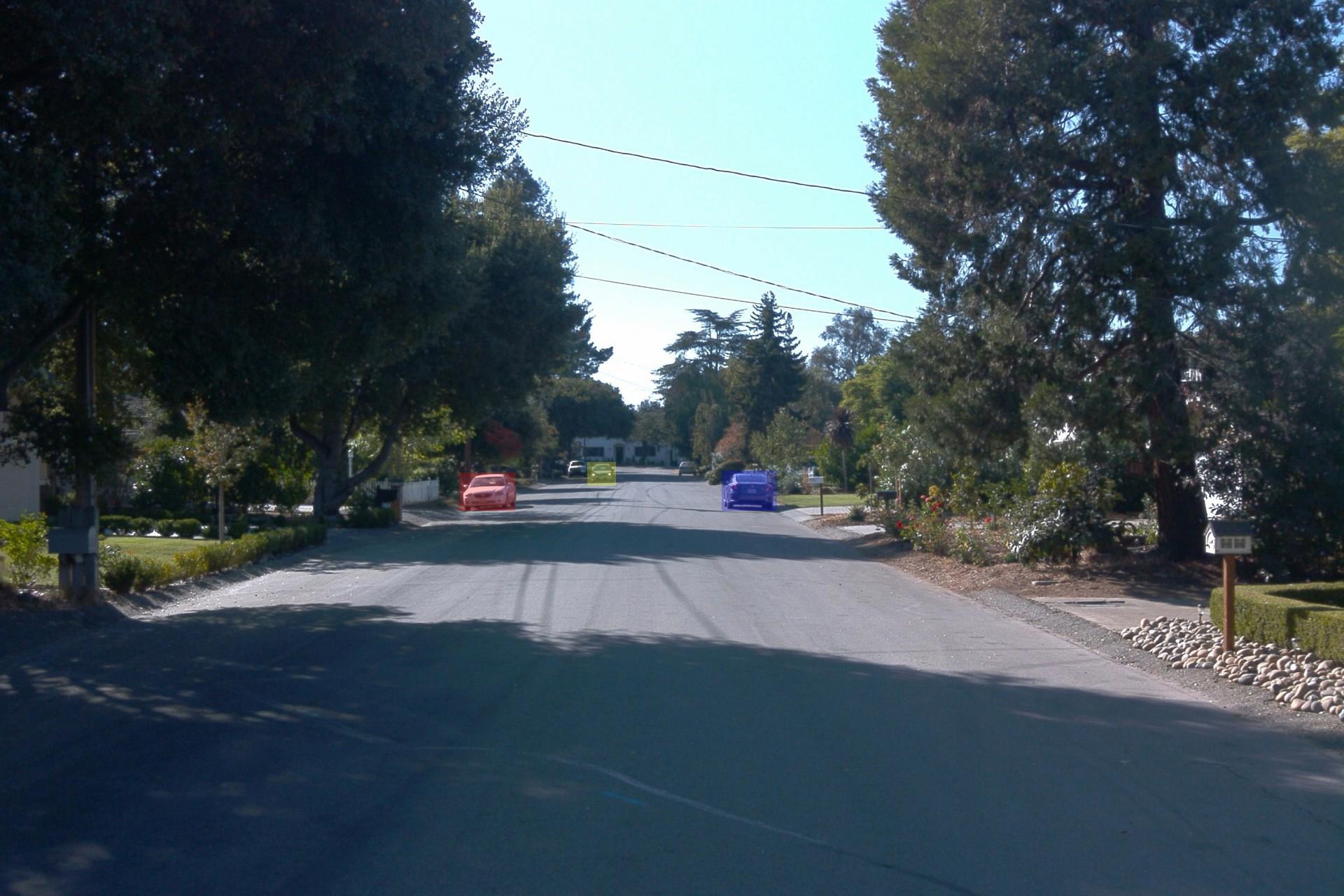}
\label{fig:vis_405_4}
\end{figure}

\begin{figure}[h]
\centering
\includegraphics[width=1\linewidth]{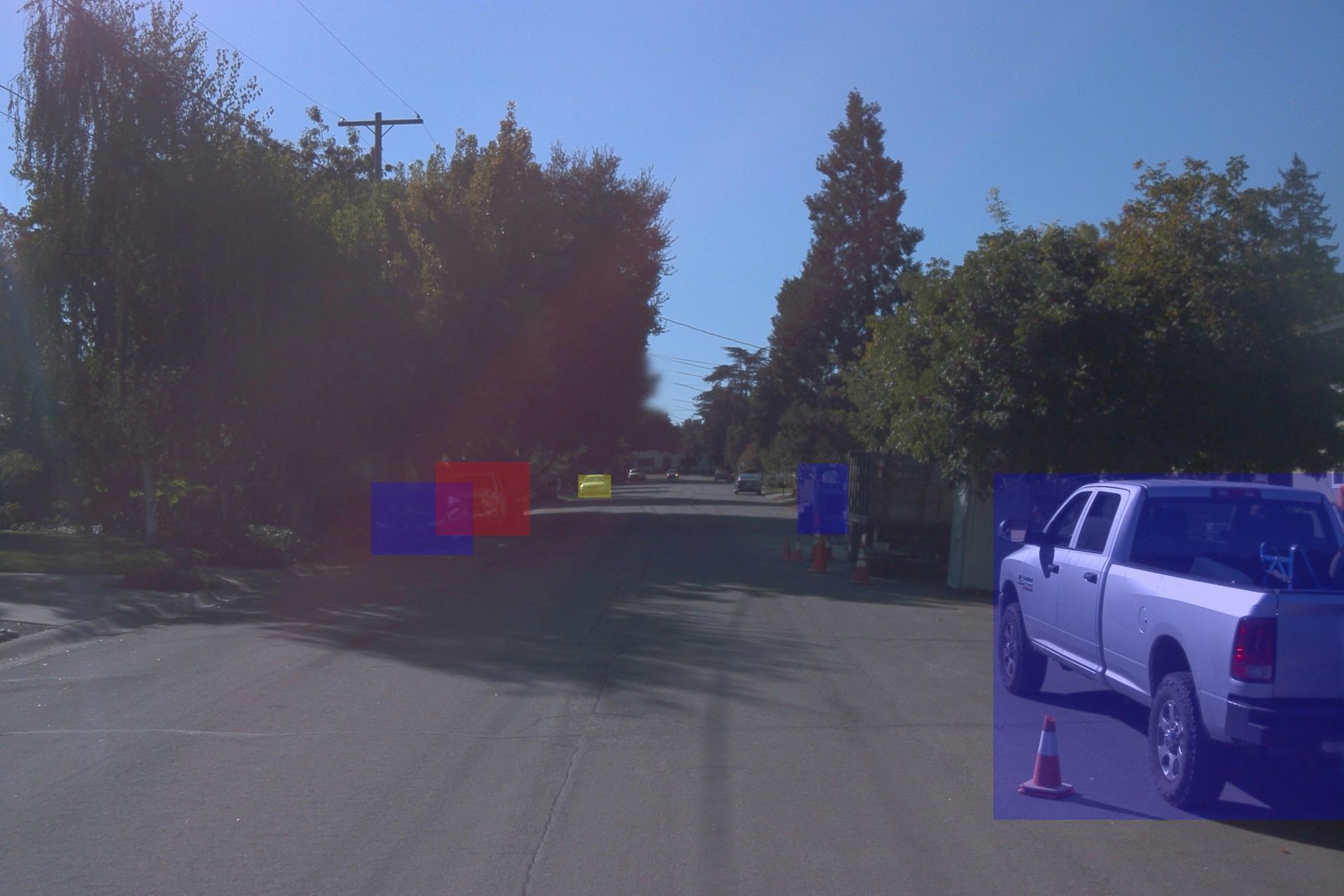}
\label{fig:vis_412_4}
\end{figure}

\begin{figure}[h]
\centering
\includegraphics[width=1\linewidth]{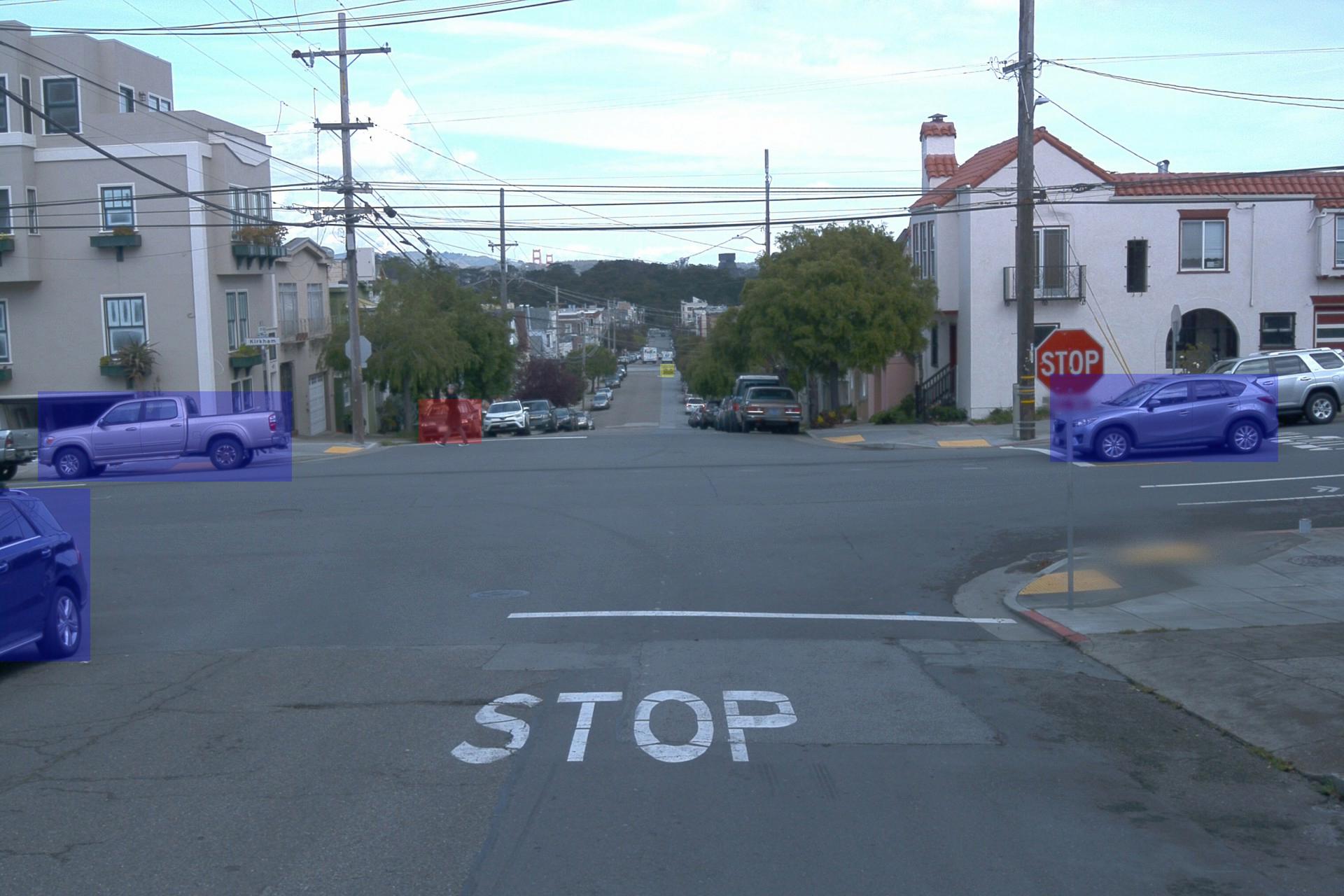}
\label{fig:vis_425_2}
\end{figure}

\begin{figure}[h]
\centering
\includegraphics[width=1\linewidth]{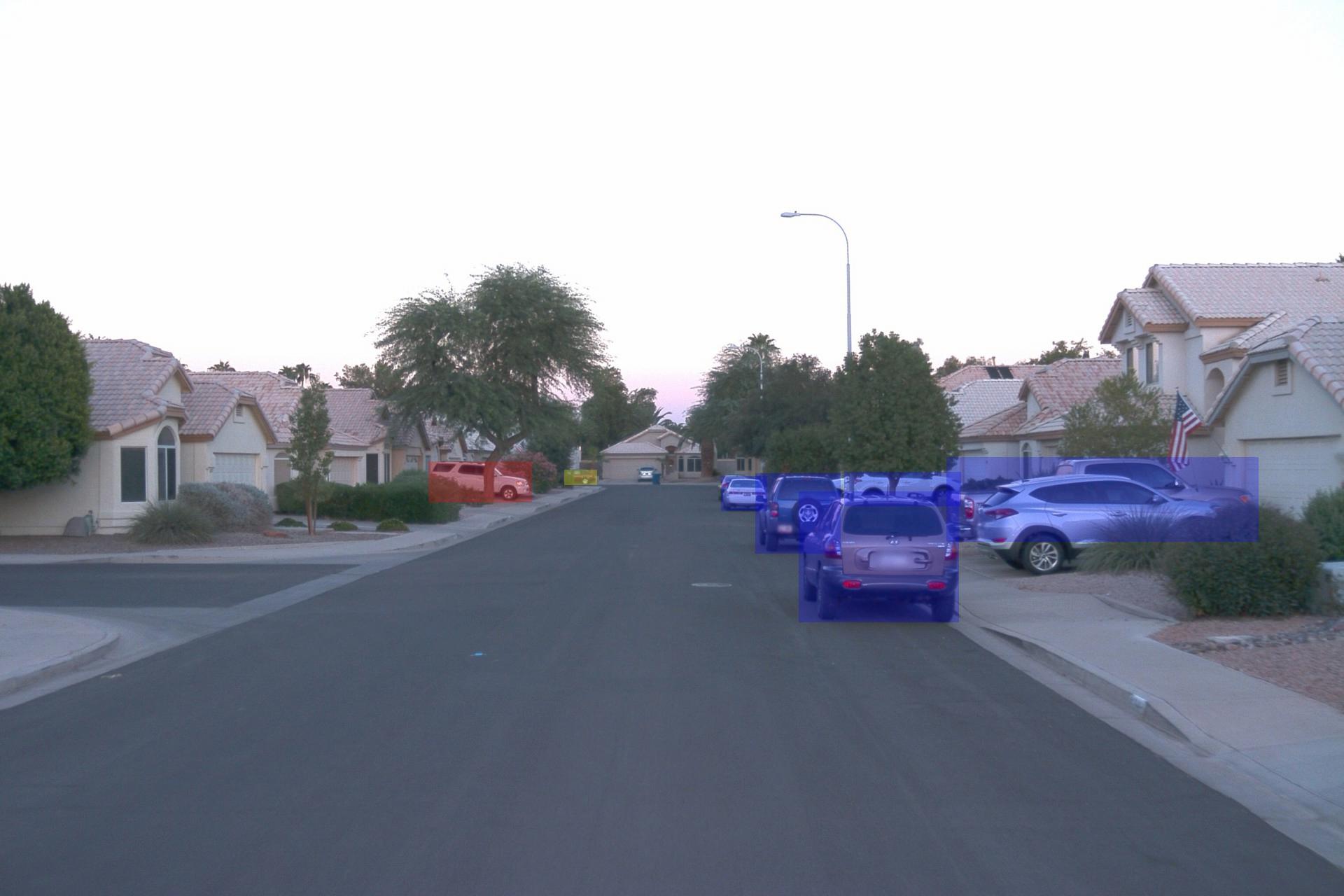}
\label{fig:vis_48_0}
\end{figure}